\documentclass{article}

\usepackage{microtype}
\usepackage{graphicx}
\usepackage{booktabs} % for professional tables

\usepackage{graphicx}
\usepackage{xcolor}
\usepackage{sidecap}
\usepackage{multirow}
\usepackage{makecell}

\usepackage{subcaption}
\usepackage{enumitem}
\usepackage[pagebackref,breaklinks,colorlinks,citecolor=cvprblue]{hyperref}

\usepackage[accepted]{icml2025}

\usepackage{amsmath}
\usepackage{amssymb}
\usepackage{mathtools}
\usepackage{amsthm}

\usepackage[capitalize,noabbrev]{cleveref}

\usepackage{placeins}  % W preambule

\theoremstyle{plain}

\theoremstyle{definition}

\theoremstyle{remark}

\AtBeginDocument{%
    \setlength\abovedisplayskip{9pt}%
    \setlength\belowdisplayskip{6pt}%
    \setlength\abovedisplayshortskip{9pt-\baselineskip}%
    \setlength\belowdisplayshortskip{6pt}
    \setlength{\itemsep}{0pt}
    \setlength{\floatsep}{11pt}
    \setlength{\textfloatsep}{11pt}}

\icmltitlerunning{How to Train Your Multi-Exit Model? Analyzing the Impact of Training Strategies}

\begin{document}

\twocolumn[
% \icmltitle{Joint or Disjoint: Mixing Training Regimes for Early-Exit Models}
% \icmltitle{Exploring the Training Strategies of Early-Exit Models}
\icmltitle{How to Train Your Multi-Exit Model?\\Analyzing the Impact of Training Strategies}
% \icmltitle{How to Train Multi-Exit Models? Analyzing the Impact of Training Strategies}



\icmlsetsymbol{equal}{*}
\icmlsetsymbol{equal_sup}{$\dagger$}

\begin{icmlauthorlist}
\icmlauthor{Piotr Kubaty}{equal,ju}
\icmlauthor{Bartosz Wójcik}{equal,equal_sup,ju}
\icmlauthor{Bartłomiej Krzepkowski}{wut}
\icmlauthor{Monika Michaluk}{uw}
\icmlauthor{Tomasz Trzcinski}{wut,toop,ideas}
\icmlauthor{Jary Pomponi}{sap}
\icmlauthor{Kamil Adamczewski}{equal_sup,wust,ideas}
\end{icmlauthorlist}

\icmlaffiliation{ju}{Jagiellonian University}
\icmlaffiliation{uw}{University of Warsaw}
\icmlaffiliation{wut}{Warsaw University of Technology}
\icmlaffiliation{toop}{Tooploox}
\icmlaffiliation{ideas}{IDEAS Research Institute}
\icmlaffiliation{sap}{Department of Information Engineering, Electronics, and Telecommunications (DIET) at Sapienza, University of Rome, Italy}
\icmlaffiliation{wust}{Wroclaw University of Science and Technology}

\icmlcorrespondingauthor{Bartosz Wójcik}{bartwojc@gmail.com}

\icmlkeywords{Machine Learning, ICML, multi-exit models, early exiting, conditional computation, dynamic neural networks, computational efficiency, efficient inference}

\vskip 0.3in
]



\printAffiliationsAndNotice{\icmlEqualContribution. $\dagger$ Equal supervision.} % otherwise use the standard text.

\begin{abstract}
Early exits enable the network's forward pass to terminate early by attaching trainable internal classifiers to the backbone network. Existing early-exit methods typically adopt either a joint training approach, where the backbone and exit heads are trained simultaneously, or a disjoint approach, where the heads are trained separately. However, the implications of this choice are often overlooked, with studies typically adopting one approach without adequate justification. This choice influences training dynamics and its impact remains largely unexplored. In this paper, we introduce a set of metrics to analyze early-exit training dynamics and guide the choice of training strategy. We demonstrate that conventionally used joint and disjoint regimes yield suboptimal performance. To address these limitations, we propose a mixed training strategy: the backbone is trained first, followed by the training of the entire multi-exit network. Through comprehensive evaluations of training strategies across various architectures, datasets, and early-exit methods, we present the strengths and weaknesses of the early exit training strategies. In particular, we show consistent improvements in performance and efficiency using the proposed mixed strategy.
\end{abstract}

\section{Introduction}

Deep neural networks have achieved remarkable results across a variety of machine learning tasks. While the depth of these networks significantly contributes to their enhanced performance, the necessity of using large models for all inputs, especially in resource-constrained environments like mobile and edge computing devices, is questionable.

\begin{figure}[t]
    \centering
    \includegraphics[scale=0.15]{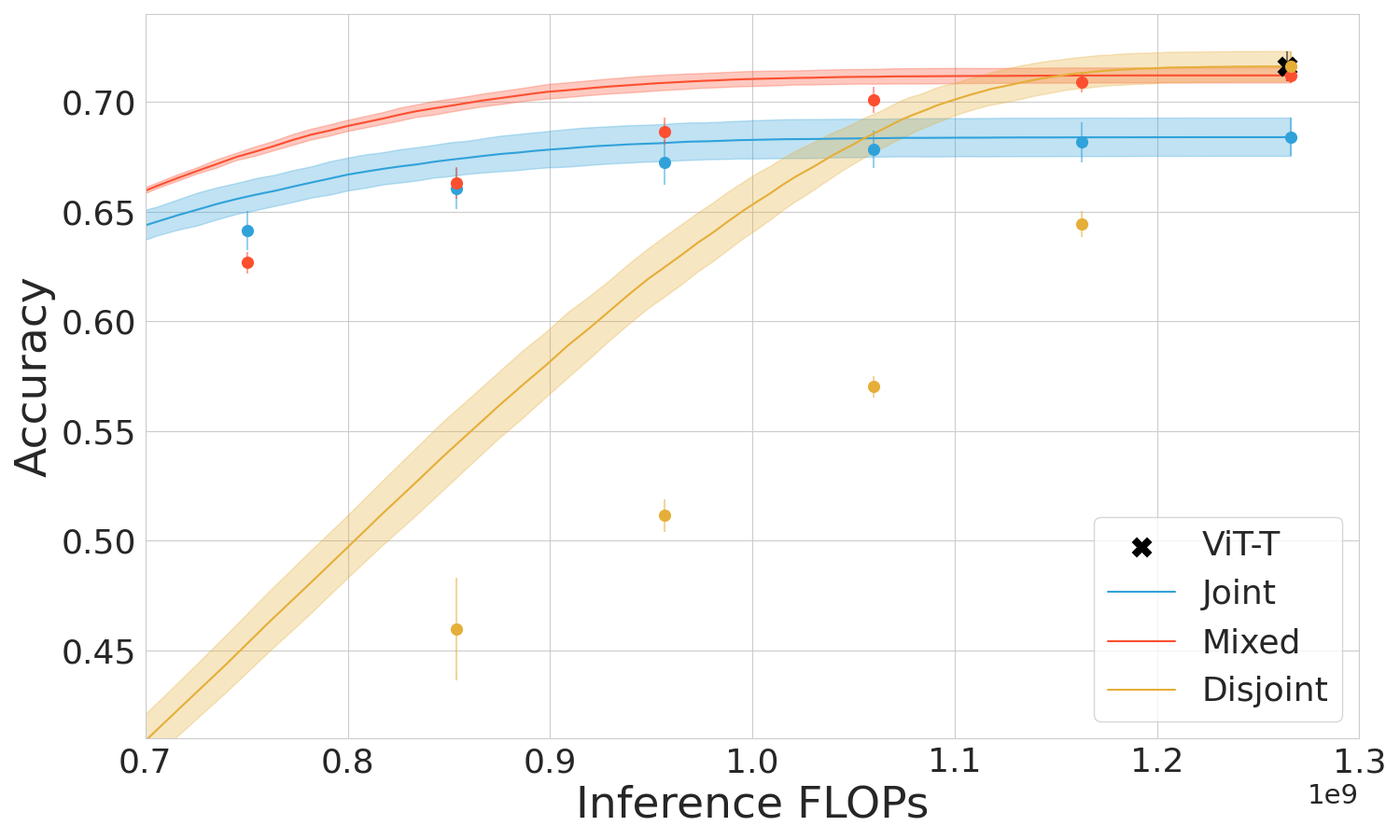}
    \caption{Performance-cost trade-off of the multi-exit ViT-T model trained on the ImageNet-1k dataset by using different training regimes. The choice of the training regime significantly impacts the performance across all computational budgets.}
    \label{fig:demo}
\end{figure}

Early exit methods for deep neural networks have gained importance due to their potential to significantly improve computational efficiency. By exiting at earlier layers, these methods can decrease the number of operations needed for computation of the forward pass, leading to faster inference times. In doing so they allow the network to adapt its computational cost to the difficulty of the input sample. Simpler inputs can be processed with fewer layers, while more complex inputs can utilize the full capacity of the network. 

Early exit methods are implemented through augmentation of the original architecture with internal classifiers (ICs) attached to selected intermediate layers \citep{kaya2019shallow}. These ICs are designed to perform classification tasks based on the representations available at their respective positions in the network. 
%Integrating intermediate classifiers into an architecture significantly alters the training dynamics and learning process. 
%However, the ways the early-exit architectures are trained are still not well understood.
A common approach for training early-exit models involves training the entire multi-exit network, including the added classifiers, from scratch \cite{huang2018multi,yang2020resolution,meronen2024fixing} (\textbf{"joint" regime}). Alternatively, some methods train the backbone network first, then freeze its weights and train the parameters of the newly added ICs in the second, separate phase of training \cite{teerapittayanon2016branchynet,liao2021global,zhou2020bert} (\textbf{"disjoint" regime}). To the best of our knowledge, no study compares or explores the relationship between these training regimes.

In this study, we perform an extensive assessment of early-exit regimes and notice the choice of training strategy has a significant impact on the final model's performance. We identify the relationship between computational budget and the choice of the regime. Using the disjoint regime results with a network that is significantly impaired when smaller computational budget is assumed. While the joint regime might initially seem as the appropriate way of training multi-exit networks, we demonstrate that due to its training dynamics it biases the network and produces a model with subpar performance on higher computational budgets. 

In order to address the weaknesses of multi-exit networks, we propose a novel \textbf{``mixed'' regime}: train the backbone network until convergence, then train the entire model jointly, including the internal classifiers, until convergence. This approach ensures that the backbone architecture is adequately trained before optimizing it alongside the internal classifiers for improved performance.

To gain a deeper understanding of learning and optimization in multi-exit architectures, we propose a set of metrics to describe the training dynamics of early-exit models trained under various regimes. Through gradient dominance metric we reveal the set of ICs that have the largest impact on the backbone during training and explain the performance gains of popular technique of loss and gradient scaling depend on the training strategy choices. Furthermore, mode connectivity allows to analyze solution similarities of different regimes. Moreover, numerical rank, and mutual information metrics aid to explain the performance of regimes under different computational budgets.

Finally, we provide a thorough empirical evaluation of early-exit regimes across different network architectures, data modalities, datasets and early-exit methods. Our results show that proposed alternative strategy enables significant improvements in performance in medium and high budgets over the commonly used joint training.

\section{Training Regimes}
\label{sec:regimes}
Early exit methods fundamentally alter the organization of neural networks. It is widely believed that neural networks develop a hierarchical representation of features, where earlier layers learn basic shapes and patterns, while later layers progressively capture more complex abstractions \cite{zeiler2014visualizing}. In other words, the earlier layers are characterized by higher frequency features while later layers learn low frequency elements. This regularity is disrupted in the case of early exit architectures as the backbone network is given additional classifiers that are placed in earlier parts of the network. These changes in architecture require a different approach for training and more nuanced analysis how the training should proceed. 

In the early-exit setting, parameters can be divided into \textit{backbone} parameters and \textit{internal classifier (IC)} parameters. Each of these two groups can be trained separately or jointly. In this paper, we frame the training process of any early-exit method as consisting of three following phases:

\textbf{Phase 1}: Train the backbone network parameters \( \theta_b \) by minimizing the loss at the final output layer (could be the last IC or an added final classifier).

\begin{equation}
    \label{eq:ph1}
    \theta_b^\ast = \arg\min_{\theta_b} \mathbb{E}_{(x_i, y_i) \sim \mathcal{D}} \left[ \mathcal{L}^{(K)}(\theta_b, \theta_{\text{IC}}^{(K)}) \right]
\end{equation}

During this phase, \( \theta_{\text{IC}} \) are either not trained or not present at all.

\textbf{Phase 2:} Train both the backbone network and the ICs simultaneously.
\begin{equation}
    \label{eq:ph2}
    \theta^\ast = \arg\min_{\theta} \sum_{k=1}^K \alpha_k \mathbb{E}_{(x_i, y_i) \sim \mathcal{D}} \left[ \mathcal{L}^{(k)}(\theta_b, \theta_{\text{IC}}^{(k)}) \right]
\end{equation}
\textbf{Phase 3:} Freeze $\theta_b^\ast$ and train only the IC parameters  $\theta_{\text{IC}}$ .
\begin{equation}
    \label{eq:ph3}
    \theta_{\text{IC}}^\ast = \arg\min_{\theta_{\text{IC}}} \sum_{k=1}^K \alpha_k \mathbb{E}_{(x_i, y_i) \sim \mathcal{D}} \left[ \mathcal{L}^{(k)}(\theta_b^\ast, \theta_{\text{IC}}^{(k)}) \right]
\end{equation}
Correspondingly, we generalize the early exit training regimes into three types based on which of the phases are used:

\noindent\textbf{Disjoint training (Phases 1+3).}
The model parameters undergo training during the first and third phases. That is, the backbone architecture is trained first, and then the ICs are trained separately while the backbone parameters are frozen.

\noindent\textbf{Joint training (Phase 2).} The training consists only of the second phase in which the entire model -- including the ICs -- is trained. It is currently the most common way of training early-exit models \cite{matsubara2022split}.  

\noindent\textbf{Mixed training (Phases 1+2).} The training consists of two phases. The backbone is trained in isolation first, and then the entire network, including the ICs, is trained jointly. The regime emphasizes the importance of backbone pre-training as a better way to initialize the architecture for further training. This is our proposed way to improve multi-exit model training.

%In Appendix we also describe a set of less popular alternative training regimes.

% In practical applications one can use a set of pre-trained weights instead of training the model from scratch. Note that all the listed regimes can be used with pre-trained weights, despite some studies claiming that the disjoint training 

\section{A Framework for Analyzing Multi-Exit Models}
\label{sec:analysis_framework}

This section proposes a framework for analysing and comparing multiple early exit models. This framework comprises multiple evaluations, each focusing on a different aspect of a model. 

\subsection{Gradient Dominance}
\label{sec:gradient_dominance}

The use of internal classifiers during training in early-exit training regime fundamentally alters the training dynamics, as these classifiers contribute to the overall loss. The gradient update now comes from multiple classifiers instead of just the final one, as in a standard neural network. 

% To study this phenomenon, we introduce a metric that helps understand which gradient is more prominent during training. We define \textit{gradient dominance} (GD) as the cosine similarity between the gradient from an individual internal classifier, $\mathbf{g}_{i}$, and the overall gradient of the model $\mathbf{g}_{\text{total}}$:

% \[
% \text{GD} = \frac{1}{ \lvert \text{IC} \rvert } \sum_{i \in \text{IC}} <\mathbf{g}_{i}, \mathbf{g}_{\text{total}}>
% \]

To study this phenomenon we calculate which gradient is more prominent during training. We define \textit{gradient dominance} (GD) as the cosine similarity between the gradient from an individual internal classifier, $\mathbf{g}_{i}$, and the overall gradient of the model $\mathbf{g}_{\text{total}}$:
\[
\text{GD}_i = <\mathbf{g}_{i}, \mathbf{g}_{\text{total}}>
\]
% 
% This leads to the following question: which gradients contribute the most to the overall gradient, and how do the gradients from different classifiers align? To answer this question, we first introduce a metric called \textit{gradient dominance}, which computes the cosine similarity between gradient from an individual internal classifier, $\mathbf{g}_{\text{IC}}$, and the overall gradient, $\mathbf{g}_{\text{total}}$:
% $$\text{gradient dominance}(\mathbf{g}_{\text{IC}}, \mathbf{g}_{\text{total}}) = \frac{\mathbf{g}_{\text{IC}} \cdot \mathbf{g}_{\text{total}}}{\|\mathbf{g}_{\text{IC}}\| \|\mathbf{g}_{\text{total}}\|}$$

where $< \cdot, \cdot>$ is the cosine similarity. Gradient Dominance measures the consistency of the gradient directions produced by the early-exit classifiers and evaluates how well gradients from separate classifiers align with the overall gradient across the entire model. If the cosine similarity is close to 1, the auxiliary classifier's gradient is highly aligned with the total gradient, indicating that it potentially dominates other ICs in its impact on the total gradient.

\subsection{Mode Connectivity}
\label{sec:mode_connectivity}

Mode connectivity theory suggests that independently trained models often exhibit similar characteristics. Notably, after training two independent models, it is possible to find a continuous path in the parameter space where the loss remains low, enabling the models to be connected without encountering high-loss regions \cite{garipov2018loss}.

Building on the observation that independently trained neural networks can be linearly connected in weight space after accounting for permutation symmetries, as described in \cite{ainsworth2022git}, we extend this idea to early-exit architectures trained under different regimes.

Instead of focusing solely on independently trained networks, we investigate early-exit architectures trained in distinct regimes. 
To align models for mode connectivity, we define an optimal permutation \( \Pi^* \) by solving $
\Pi^* = \arg\min_{\Pi} \| \Theta_A - \Pi \Theta_B \|_F$
 where \( \| \cdot \|_F \) is the Frobenius norm, and $ \Theta_A$ and $ \Theta_B$ are the parameters of two early-exit models trained under different strategies as described in Sec.~\ref{sec:regimes}. Given the optimal permutation, the permuted interpolation path is:

\begin{equation}
\Theta(\lambda) = (1 - \lambda) \Theta_A + \lambda \Pi^* \Theta_B
\end{equation}

The core idea is based on the fact that, if two models trained with different strategy can be interpolated such that the resulting model has low loss over the whole interpolation path, then the regime solutions lie in the similar loss basin.

\subsection{Numerical Rank}
\label{sec:numerical_rank}

So far, the proposed tools analyse the gradients of the model or their final predictions. Here, we propose to evaluate 
the expressiveness of a model by analysing its intermediate activations, using the numerical ranks of activation maps \citep{masarczyk2024tunnel}. Mathematically, for a given layer $i$, the rank is evaluated as:
\begin{equation}
    r_i = \text{Rank}(A_i), \quad A_i \in \mathbb{R}^{n \times m}
\end{equation}
where \( A_i \) is the activation matrix of dimensions \( n \) (number of samples) and \( m \) (number of features).

The rank of the internal representations associated with different layers can provide insight into the ``expressiveness'', or capacity of the network. A higher rank (closer to the maximum possible for a given layer’s matrix dimensions) indicates that the layer can capture more complex patterns or features in the data, as it implies a greater degree of linear independence among the feature detectors in that layer. High-rank activations matrices in a network suggest that the network is utilizing its capacity to learn diverse, high-frequency features, whereas a low rank might indicate that the network is not fully exploiting its potential. 

\begin{figure*}
    \centering
    \begin{subfigure}{0.32\textwidth}
        \centering
        \includegraphics[width=\linewidth]{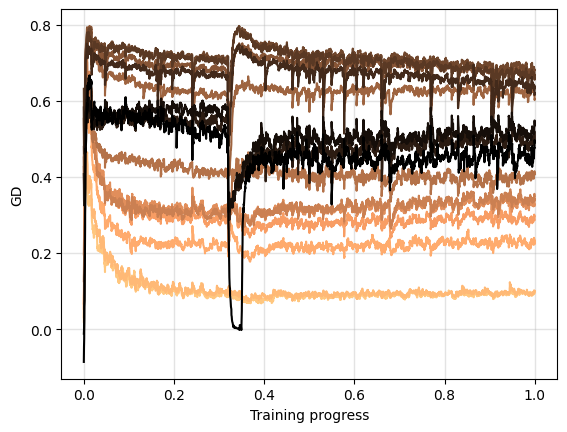}
        \caption{Joint regime}
    \end{subfigure}
    \hfill
    \begin{subfigure}{0.32\textwidth}
        \centering
        \includegraphics[width=\linewidth]{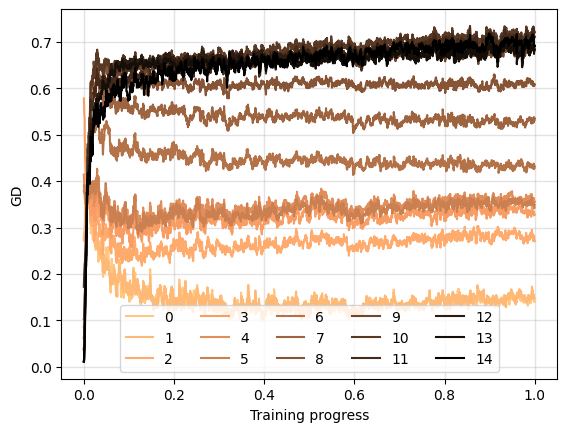}
        \caption{Mixed regime}
    \end{subfigure}
    \hfill
    \begin{subfigure}{0.32\textwidth}
        \centering
        \includegraphics[width=\linewidth]{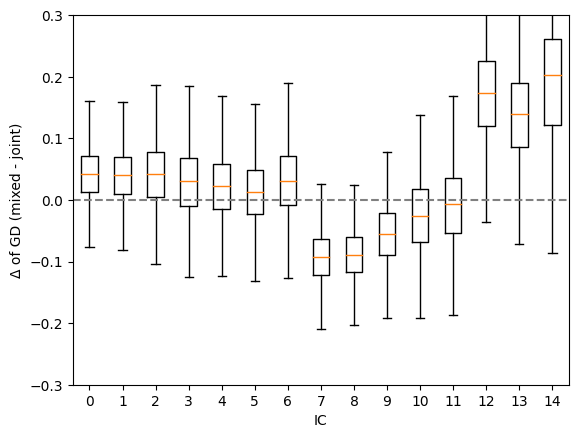}
        \caption{Avg. difference between joint and mixed}
    \end{subfigure}
    \caption{Gradient dominance for different regimes. Each line indicates how well gradients from different ICs align with the total gradient over the course of the training. The last IC dominates the most in the mixed regime, which explains its excellent performance on higher computational budgets (ResNet-50, Tinyimagenet). }
    \label{fig:gradient-dominance}
\end{figure*}

\subsection{Mutual Information}

Adding intermediate early exits to a model could alter the information flow within the model itself. To study how much such exits affect this aspect, we use the concept of mutual information. 
%Mutual information (MI) is a measure of the amount of information that one random variable contains about another random variable. 
In the context of neural networks, the mutual information between \(X\) and \(Z\) represents how much information the input \(X\) provides about the internal representation \(Z\) after passing through a neural network.  For random variables \( X \) and \( Z \), the mutual information is defined as:
$   
    I(X; Z) = \int_{x \in \mathcal{X}} \int_{z \in \mathcal{Z}} p(x, z) \log \frac{p(x, z)}{p(x)p(z)} \, dx \, dz$   
where \( p(x, z) \) is the joint probability distribution of \( X \) and \( Z \), and \( p(x) \) and \( p(z) \) are the marginal distributions of \( X \) and \( Z \), respectively. In practical terms, for neural networks, we use Monte Carlo sampling to estimate \( I(X; Z) \) due to the high dimensionality of feature spaces, and as such the results are not always monotonically decreasing~\citep{shwartz2017opening}.

In their work, \cite{kawaguchi2023does} utilize the concept of mutual information between \(X\) and \(Z\) to study the information bottleneck (IB) principle. It aims to find a balance between the information carried by \(Z\) in predicting the target \(Y\) and its complexity in terms of its mutual information with the input \(X\). Specifically, minimizing \(I(X; Z)\) reduces the complexity and overfitting by ensuring \(Z\) retains only the essential information from \(X\), and maximizing \(I(Y; Z)\) ensures that the representation \(Z\) is informative enough to predict the target variable \(Y\) effectively. To visually show how mutual information is affected by the training regime, we calculate the mutual information for each layer averaged over multiple inputs. 

% $$
% \text{MI}_z = I(X; z)
% $$

% for all layers' internal features $z$

% in which we estimate the mutual information by sampling multiple internal features from the set $\mathcal{Z}$ containing all of them. 

% \subsection{Cross similarity}
% Having multiple exits all trained on the same task could lead multiple ones to produce similar results. Here, we present a simple metric showing how much each exit is similar to others. Relating such metrics to others could give a deeper insight into how training models have early exits. The metric measures the average distance between pairs of exits as:

% $$
% \text{CS}(x) = \frac{1}{\lvert \text{IC} \rvert^2} \sum_{i \in \text{IC}} \sum_{j \in \text{IC}} <f_{i}(x), f_{j}(x) > 
% $$ 

% \textcolor{red}{the formulation must be improved} where $f_{i}$ is the output of the $i$-th layer, and $<\cdot,\cdot>$ the cosine similarity. 

\section{Empirical Evaluation of Training Regimes}

\subsection{Experimental Setup}
In this section, we outline the setup for our empirical experiments. We release the source code of our experiments at: \url{https://github.com/kamadforge/early-exit-benchmark}. A more detailed description can be found in~\Cref{sec:training_details}.

% \subsection{Experimental set-up}

% \noindent\textbf{Experimental set-up.} In this section, we perform tests on the commonly used simple early-exit method SDN \cite{kaya2019shallow}, in which sample exits early if the confidence of a classifier is larger than a predefined threshold. We also include, MSDNet -- a convolutional neural network architecture designed specifically for multi-exit models \citep{huang2018multi}. In the next section we test regimes on a range of early-exit methods. To ensure proper model training, we test different learning rates for pre-training the backbone and separately for the next phase of training in each regime, and select the optimal one for each phase. %This way we ensure that the model is trained properly. 
% For proper model convergence in each phase we utilize early-stopping to select a termination point of the training.
% %we derive a common validation set from the training set to facilitate early stopping. 
% This procedure halts training if there is no improvement in validation set accuracy over a specified number of epochs. 
% % At that point, we revert to the model checkpoint with the highest validation set accuracy.
% In appendix we include all the experimental details for better reproducibility.

\noindent\textbf{Architectures and datasets.} We conduct experiments across various datasets spanning computer vision (CV) and natural language processing (NLP). For CV, we utilize CIFAR-100 \cite{cifar}, ImageNet-1k \cite{ILSVRC15}, TinyImageNet \cite{Le2015TinyIV}, and Imagenette \cite{imagenette}. For NLP, we evaluate on 20-Newsgroups \cite{lang1995newsgroups} and STS-B \cite{wang-etal-2018-glue} datasets. As far as vision architectures are concerned, we evaluate ResNet \cite{he2016deep} and Vision Transformers (ViT) \cite{alexey2020image}. Additionally, we explore MSDNet \cite{huang2018multi}, an architecture dedicated for multi-exit models. For NLP tasks, we utilize BERT \cite{devlin2018bert}. Unless otherwise specified, the Shallow-Deep Network (SDN) \cite{kaya2019shallow} early exit model is implemented on top of a chosen backbone.

% TODO Piotr
% please add a citation for each architecture and dataset that you mention
% mention that MSDNet is an architecture explicitly designed for multi-exit models
% keep it concise, do not add unnecessary details about the datasets or architectures - the entire paragraph should be 3-6 sentences
% We conduct experiments on computer vision (CV) datasets: CIFAR-100 \cite{cifar}, ImageNet-1k \cite{ILSVRC15},  TinyImageNet \cite{Le2015TinyIV}, Imagenette \cite{imagenette}; as well as natural language processing datasets: 20 Newsgroups \cite{lang1995newsgroups}, STS-B \cite{wang-etal-2018-glue}. We consider standard architectures for CV, such as ResNets \cite{he2016deep} and Vision Transformers \cite{alexey2020image}. Furthermore, we consider MSDNet \cite{huang2018multi}, an architecture explicitly designed for multi-exit models. For NLP, we employ \cite{devlin2018bert}. Unless otherwise stated, the SDN \cite{kaya2019shallow} early exit model is built upon backbone network.
% In addition, we consider PBEE \cite{zhou2020bert} and GPF \cite{liao2021global}

% TODO Piotr
% Mention:
% -optimizer
% -LR scheduling
% -early stopping 
\noindent\textbf{Training set-up.}\label{par:exp_setup-train_setup}
We train our models, employing the AdamW optimizer \cite{loshchilov2019decoupled} alongside the Cosine Annealing scheduler with warm restarts. To ensure fair convergence across different regimes, we incorporate an early stopping mechanism. Training is terminated only when, over $n$ consecutive epochs, none of the exits achieve an improved performance compared to their best scores recorded thus far. These scores -- accuracy for classification tasks and loss for regression tasks -- are evaluated on a dedicated early-stopping validation set. All results in this section are averaged over three runs, each one having a different initial seed.

% We train our models using the PyTorch library \cite{paszke2019pytorch}, with the AdamW optimizer \cite{loshchilov2019decoupled} and the CosineAnnealingWarmRestarts scheduler. To ensure convergence, we implement an early stopping mechanism. For each experiment, a fixed number of $n$ epochs is chosen. The training phase halts only if, for $n$ consecutive epochs, none of the exits achieve a better score (i.e., accuracy for classification tasks and loss for regression tasks) than its best score achieved so far. The score is computed on the early-stopping validation set.

\noindent\textbf{Evaluation protocol.} The core of our analysis is the framework proposed to understand how adding early exits affects the model's internal state and hence the predicted values. To this end, in \Cref{sec:framework_eval}, we use the proposed framework to compare all the training regimes. Additionally, following the core idea that a model having early exits must be used to spare computational power, we evaluate all the resulting models under the lens of FLOPs saved and the obtained accuracy. These results are presented in~\Cref{sec:numerical_eval}.

\subsection{Framework Evaluation}
\label{sec:framework_eval}

In this section, we evaluate the training regimes through the framework proposed in \Cref{sec:analysis_framework}.

\noindent\textbf{Gradient Dominance.} In~\Cref{fig:gradient-dominance}, we present the gradient dominance results for the joint and mixed regimes, highlighting their distinct training dynamics. In the joint regime, optimization focuses on subnetworks in the middle of the architecture, where gradients from intermediate classifiers exert the strongest influence. Conversely, in the mixed regime, gradients are dominated by deeper classifiers, causing the early layers to primarily support the learning objectives of later classifiers, potentially at the expense of earlier ones.

The dominance of the final intermediate classifier (IC) in the mixed regime suggests its suitability for scenarios with higher computational budgets, which is supported by our empirical results. This behavior mirrors the training of a backbone, where the model is primarily guided by the loss from the final classifier. Consequently, the mixed regime tends to produce models that closely resemble standard neural networks, a fact we confirm in the next section through mode connectivity analysis.

%Note that this observation is in line and with the study of information flow and explains the results of our empirical evaluation. The fact that the gradient of the last IC dominates the total gradient during almost the entire training period in the mixed regime explains why it never leaves the basin that models trained with the disjoint regime occupy.

\begin{figure}
    \centering
    \includegraphics[width=0.8\linewidth]{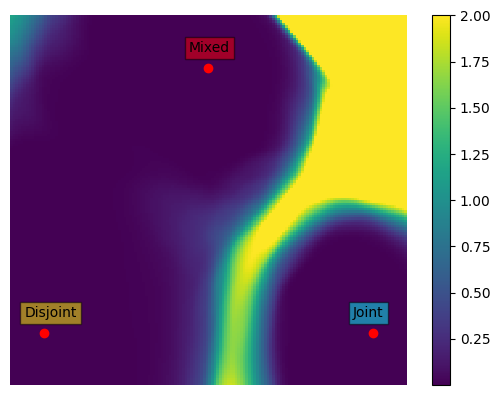}
    \caption{Mode connectivity between models trained with different training regimes . Colors represent the values of loss function, with yellow representing high loss ($\geq$ 2.0). Disjoint and mixed regimes produce similar models, while the model trained in joint regime lies in a different basin (ResNet-20, CIFAR-10).}
    \label{fig:mode_connectivity}
\end{figure}

\noindent\textbf{Mode Connectivity.} After accounting for permutation symmetries, the loss remains low during linear interpolation of the weights of models trained in the mixed and disjoint regimes, as shown in the mode connectivity plot in \Cref{fig:mode_connectivity}. This indicates that the resulting weights lie in the same basin, and the two regimes produce similar solutions. However, the disjoint regime is more constrained because it trains only the internal classifiers during the second phase. In contrast, the mixed regime updates the backbone to accommodate the added ICs, resulting in lower overall loss.

Joint training, on the other hand, leads to a model in a different basin of the loss landscape. This suggests that training the entire model at once, without "pre-trained" backbones, produces solutions that differ significantly in structure and performance.

\begin{figure}[t]
    \centering
    \includegraphics[width=\linewidth, trim=0 0 0 25, clip]{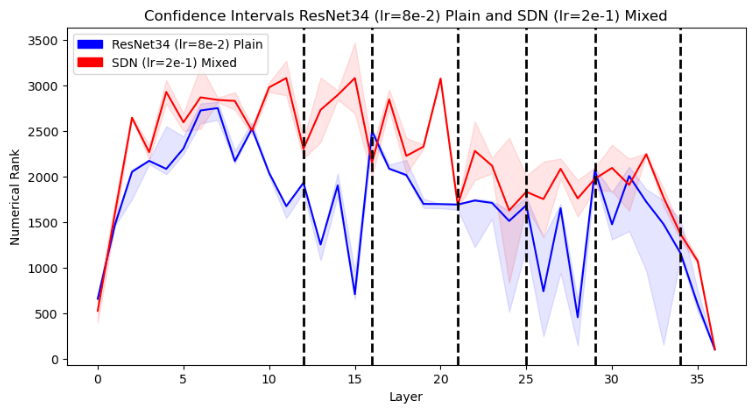}
    % First row
    %     \begin{subfigure}{0.49\textwidth} % Adjusted width to accommodate 2 images per row
    %         \includegraphics[width=\linewidth, trim=0 0 0 25, clip]{figs/numrank_phase1vsphase2.png}
    %         \caption{The change in expressiveness of the network from Phase 1 (backbone) to Phase 2 (backbone+ICs).}
    %     \label{fig:phase1phase2_n}
    % \end{subfigure}
    % \hfill
    % \begin{subfigure}{0.49\textwidth}
    %     \includegraphics[width=\linewidth, trim=0 0 0 25, clip]{figs/numrank_regimes.png}
    %     \caption{Mixed vs joint training. The vertical lines indicate IC placement.}
    %     \label{fig:numrank_mixedjoint}
    % \end{subfigure}
    \caption{Numerical ranks as measured in the backbone network of the multi-exit model which was trained with different regimes. The rank increases when the ICs are allowed to affect the backbone network.}
    \label{fig:numerical_rank}
\end{figure}

\noindent\textbf{Numerical Rank.} In our framework, we analyze the numerical rank of the backbone model under different early-exit regimes. A regular neural network typically exhibits a higher rank in earlier layers and a lower rank in deeper layers, as illustrated in~\Cref{fig:numerical_rank}. Note that training only the backbone corresponds to the model obtained in the disjoint regime, as this training strategy does not modify the backbone.

We observe a distinct change in network expressiveness once intermediate classifiers are attached and permitted to influence the backbone. The numerical rank rises across the layers, becoming comparable to the numerical rank of a multi-exit network trained entirely from scratch. This suggests that well-functioning multi-exit models require layers with greater expressiveness, which underpins the effective operation of all subsequent internal classifiers. Such an increase in expressiveness is impeded when employing the disjoint training strategy.

% ~\Cref{fig:numrank_mixedjoint} shows the difference between the network trained with mixed and joint regime. The mixed regime has a flatter structure with relatively lower ranks earlier and higher ranks later. We hypothesize this since early-exit architecture consists of a set of classifiers placed across the network, the flatter architecture is desired for better performance across all the classifiers. In contrast, a steeper curve as in the case of joint regime may resemble more a regular architecture and be less suitable for early-exit task.

\begin{figure}[b]
    \centering
    \includegraphics[width=\linewidth, height=4cm, trim=0 0 0 30, clip]{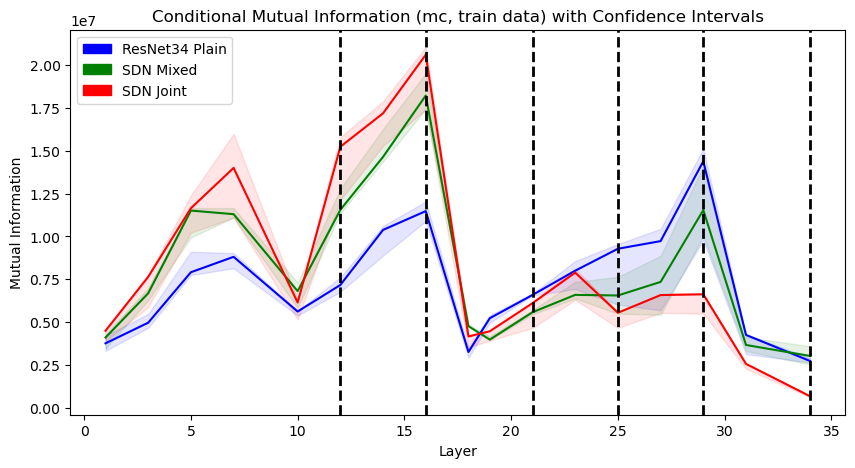}
    \caption{Mutual information $I(X;Z)$ between the input $X$ and the internal representation $Z$ of the backbone at different depths for the three multi-exit model training regimes.}
    \label{fig:mutual_information}
\end{figure}

\noindent\textbf{Mutual Information.} Early-exit architectures attach intermediate classifiers (ICs) to internal layers, altering the distribution of information flow as seen in ~\Cref{fig:mutual_information}. The effect is two-fold and differs between earlier and later layers. 
\textit{Earlier layers:} The mutual information between $X$ and $Z$ is \textit{larger} compared to a network trained without additional classifiers. 
\textit{Deeper layers:} The mutual information for early-exit architecture is \textit{lower} in the final layers. 
% We hypothesize that exiting earlier makes the information about the input less necessary in deeper layers since the decision is often made in the earlier parts of the architecture.

The above effect is seen in both regimes but is more pronounced in the joint regime. The information flow in the joint regime is more skewed and different from backbone-only training. Backbone training in the mixed regime makes the information flow fall between backbone-only and joint training.
This is due to the fact that the representation of easy samples is not complex (that is, it is processed with just a few layers before exiting through an early IC). As the sample is easy, it is clearly and distinctly located within the boundaries of a single class.  
To describe it in terms of mutual information, the network does not need to reduce the complexity of $X$ to fit the internal representation $Z$, as $X$ has little irrelevant details.
Consequently, the input $X$ is not compressed and the internal representation $Z$ has similar complexity to the representation of $X$, hence $I(X;Z)$ is higher.

Following this observation, we note that with higher $I(X;Z)$ in earlier layers, the joint strategy is more suitable for easy datasets where more samples exit at earlier layers. 
Similarly, the mixed regime learns more uniform representation of the $I(X;Z)$ across the network (one may observe an analogy to the numerical rank results in~\Cref{sec:numerical_rank}) and may be preferred for more difficult datasets that exit at later internal classifiers.

\begin{table}[t]
    \centering
    \caption{Testset accuracy of multi-exit models trained with different regimes on the CV datasets. Each column represents the maximum averaged computational budget assumed for the model, indicated as a percentage of the computational cost of the backbone. The mixed regime achieves significant improvements over the commonly used joint and disjoint regimes.}
    \label{tab:main_results}
    \begin{tabular}{cccccc}
        \toprule
        Setting & Regime & 25\% & 50\% & 75\% & 100\% \\
        \midrule
        \multirowcell{3}{ResNet-34\\C-100} & Disjoint & $52.57$ & $67.56$ & $73.49$ & $73.79$ \\
        & Joint & $\textbf{62.84}$ & $72.80$ & $74.32$ & $74.17$ \\
        & Mixed & $62.24$ & $\textbf{73.81}$ & $\textbf{75.92}$ & $\textbf{75.88}$ \\
        \midrule
        \multirowcell{3}{MSDNet\\C-100} & Disjoint & $56.74$ & $63.96$ & $68.59$ & $70.36$\\
        & Joint & $65.93$ & $\textbf{72.02}$ & $74.73$ & $75.86$ \\
        & Mixed & $\textbf{65.94}$ & $72.01$ & $\textbf{75.46}$ & $\textbf{76.51}$ \\
        \midrule
        \multirowcell{3}{ViT-T\\C-100} & Disjoint & $28.33$ & $47.32$ & $61.50$ & $63.99$ \\
        & Joint & $40.87$ & $60.48$ & $66.22$ & $66.49$ \\
        & Mixed & $\textbf{41.65}$ & $\textbf{64.07}$ & $\textbf{70.09}$ & $\textbf{70.25}$ \\
        \midrule
        \multirowcell{3}{ResNet-50\\Tiny-IN} & Disjoint & $38.49$ & $49.25$ & $60.50$ & $65.71$ \\
        & Joint & $\textbf{52.98}$ & $62.16$ & $65.14$ & $65.01$ \\
        & Mixed & $52.89$ & $\textbf{63.28}$ & $\textbf{67.20}$ & $\textbf{67.24}$ \\
        \midrule
        \multirowcell{3}{ViT-T\\IN-1k} & Disjoint & $10.22$ & $35.15$ & $61.23$ & $\textbf{71.61}$ \\
        & Joint & $\textbf{36.39}$ & $61.89$ & $68.08$ & $68.39$ \\
        & Mixed & $35.91$ & $\textbf{62.96}$ & $\textbf{70.79}$ & $71.20$ \\
        \midrule
        \multirowcell{3}{ViT-S\\IN-1k} & Disjoint & $10.23$ & $33.91 	$ & $68.02$ & $\textbf{78.38}$ \\
        & Joint & $\textbf{50.10}$ & $73.99$ & $76.38$ & $76.44$ \\
        & Mixed & $49.50$ & $\textbf{75.17}$ & $\textbf{78.28}$ & $78.33$ \\
        \bottomrule
    \end{tabular}
\end{table}

% \begin{table}[h!]
%     \centering
%     \caption{Testset accuracy of multi-exit models trained using different regimes on the ImageNet-1k dataset. The mixed regime effectively closes most of the performance gap observed in the joint regime at higher computational budgets.}
%     \label{tab:imagenet}
%     \begin{tabular}{cccccc}
%         \toprule
%         Backbone & Regime & 25\% & 50\% & 75\% & 100\% \\
%         \midrule
%         \multirow{3}{*}{ViT-T} & Disjoint & $10.22$ & $35.15$ & $61.23$ & $71.61$ \\
%         & Joint & $36.39$ & $61.89$ & $68.08$ & $68.39$ \\
%         & Mixed & $35.91$ & $62.96$ & $70.79$ & $71.20$ \\
%         \bottomrule
%     \end{tabular}
% \end{table}

\subsection{Performance-Cost Evaluation}
\label{sec:numerical_eval}

\Cref{tab:main_results} presents the performance of the training regimes on various dataset and architectures. The \textbf{disjoint regime shows weaker performance under lower computational budgets}, as freezing the backbone limits the ability to learn effective early-layer representations for classification. The results for ImageNet-1k, which are also reported in ~\Cref{fig:demo}, demonstrate a similar trend, but also highlighting that the \textbf{joint regime experiences a performance gap at higher computational budgets}. The mixed regime performs consistently well across all scenarios. Due to space constraints in the main paper we present the result averaged over three runs, but in~\Cref{sec:full_results} we present these results with standard deviations. In~\Cref{sec:alt_regimes} we also examine four alternative training regimes.

% \paragraph{NLP \& Regression}
\paragraph{Generalization to Natural Language Processing \& Regression Tasks.} To explore the generalizability of these results, we first conduct similar experiments on natural language processing data and on a regression task, with the results presented in~\Cref{tab:nlp_results}. These results are consistent with those observed before, further highlighting the weaknesses of both the joint and disjoint regimes.

\begin{table}[t]
    \centering
    \caption{Testset accuracy (classification) and MSE (regression) of multi-exit  BERT-B models trained using different regimes on the NLP datasets. The results align with those observed for classification tasks.}
    \label{tab:nlp_results}
    \begin{tabular}{cccccc}
        \toprule
        Dataset & Regime & 25\% & 50\% & 75\% & 100\% \\
        \midrule
        \multirowcell{3}{Newsgr.\\(class.)} & Disjoint & $68.53$ & $83.75$ & $\textbf{85.75}$ & $\textbf{85.69}$ \\
        & Joint & $84.24$ & $84.41$ & $84.41$ & $84.41$ \\
        & Mixed & $\textbf{84.99}$ & $\textbf{85.25}$ & $85.25$ & $85.25$ \\
        \midrule
        \multirowcell{3}{STS-B\\(regr.)} & Disjoint & $2.37$ & $1.55$ & $0.54$ & $\textbf{0.51}$ \\
        & Joint & $\textbf{1.70}$ & $0.61$ & $\textbf{0.53}$ & $0.52$ \\
        & Mixed & $2.43$ & $\textbf{0.59}$ & $\textbf{0.53}$ & $\textbf{0.51}$ \\
        \bottomrule
    \end{tabular}
\end{table}

% \begin{table}
%     \centering
%     \begin{tabular}{cccccc}
%         \toprule
%         Backbone & Regime & 25\% & 50\% & 75\% & 100\% \\
%         \midrule
%         \multirow{3}{*}{BERT-B} & Disjoint & $68.53$ & $83.75$ & $85.75$ & $85.69$ \\
%         & Joint & $84.24$ & $84.41$ & $84.41$ & $84.41$ \\
%         & Mixed & $84.99$ & $85.25$ & $85.25$ & $85.25$ \\
%         \bottomrule
%     \end{tabular}
%     \caption{Testset accuracy of multi-exit models trained using different regimes on the 20 Newsgroups dataset. Similar to the results on computer vision datasets, the mixed regime consistently outperforms the joint regime.}
%     \label{tab:newsgroups}
% \end{table}

% \begin{table}
%     \centering
%     \begin{tabular}{cccccc}
%         \toprule
%         Backbone & Regime & 25\% & 50\% & 75\% & 100\% \\
%         \midrule
%         \multirow{3}{*}{BERT-B} & Disjoint & $2.37$ & $1.55$ & $0.54$ & $0.51$ \\
%         & Joint & $1.70$ & $0.61$ & $0.53$ & $0.52$ \\
%         & Mixed & $2.43$ & $0.59$ & $0.53$ & $0.51$ \\
%         \bottomrule
%     \end{tabular}
%     \caption{Testset mean squared error of multi-exit models trained using different regimes on the STS-B dataset. The results align with those observed for classification tasks.}
%     \label{tab:stsb}
% \end{table}

\paragraph{Generalization to Other Early-Exit Methods.}
To show that our conclusions extend to other early-exit approaches, we conduct three additional experiments for different early-exit methods: GPF \cite{liao2021global}, PBEE \cite{zhou2020bert}, and one that uses entropy instead of max-softmax probability as a proxy for confidence~\cite{teerapittayanon2016branchynet}. The findings from these experiments are consistent with our main results, again showing the effectiveness of the mixed regime.

\begin{table}[t]
    \centering
    \caption{CIFAR-100 testset accuracy of multi-exit models trained with different regimes on the following early-exit methods: GPF \cite{liao2021global}, PBEE \cite{zhou2020bert}, and entropy-based exit criterion \cite{teerapittayanon2016branchynet}. The results are consistent with the main experiments.}
    \begin{tabular}{cccccc}
        \toprule
        EE Method & Regime & 25\% & 50\% & 75\% & 100\% \\
        \midrule
        \multirow{3}{*}{PBEE} & Disjoint & $19.28$ & $44.11$ & $57.85$ & $63.91$ \\
        & Joint & $28.66$ & $56.49$ & $64.31$ & $66.52$ \\
        & Mixed & $\textbf{28.87}$ & $\textbf{59.42}$ & $\textbf{68.33}$ & $\textbf{70.33}$ \\
        \midrule
        \multirow{3}{*}{GPF} & Disjoint & $33.16$ & $53.19$ & $62.54$ & $64.01$ \\
        & Joint & $46.21$ & $62.23$ & $66.84$ & $67.19$ \\
        & Mixed & $\textbf{47.21}$ & $\textbf{64.51}$ & $\textbf{68.92}$ & $\textbf{69.10}$ \\
        \midrule
        \multirow{3}{*}{Entropy} & Disjoint & $27.14$ & $46.09$ & $60.30$ & $63.99$ \\
        & Joint & $41.16$ & $58.60$ & $65.69$ & $66.49$ \\
        & Mixed & $\textbf{41.54}$ & $\textbf{62.20}$ & $\textbf{69.63}$ & $\textbf{70.25}$ \\
        \bottomrule
    \end{tabular}
    \label{tab:other_methods}
\end{table}

\paragraph{Generalization to the Pre-trained Setup.} Nowadays, most applications begin with a model pre-trained on a large dataset, which is then fine-tuned for a specific target task. We emphasize that during the fine-tuning step, any of the multi-exit training regimes can be applied. To assess whether our findings extend to this scenario, we use the weights of a pre-trained ViT-B model from the \emph{torchvision} library and fine-tune it on CIFAR-100 using the three training regimes. The results, presented in \Cref{tab:transfer_learning_cifar100}, demonstrate patterns similar to those observed in the from-scratch training setup. Specifically: (1) the disjoint regime continues to show significantly lower performance under lower computational budgets, and (2) the joint regime still exhibits a performance gap at higher computational budgets.

\begin{table}
    \centering
    \caption{Test accuracy of a ViT-B model pre-trained on ImageNet-1k and fine-tuned on CIFAR-100. The disjoint regime struggles at low budgets, while the joint regime shows a slight but notable performance gap at higher budgets. The mixed regime performs consistently well, aligning with the patterns observed in the main experiments.}
    \begin{tabular}{ccccc}
        \toprule
        Regime & 35\% & 50\% & 75\% & 100\% \\
        \midrule
        Disjoint & $28.50$ & $55.64$ & $87.49$ & $\textbf{89.95}$ \\
        Joint & $73.18$ & $85.41$ & $87.82$ & $87.85$ \\
        Mixed & $\textbf{73.38}$ & $\textbf{85.70}$ & $\textbf{88.06}$ & $88.23$ \\
        \bottomrule
    \end{tabular}
    \label{tab:transfer_learning_cifar100}
\end{table}

\subsection{Impact of Gradient and Loss Scaling}
The gradient dominance results demonstrated how the mixed regime prioritizes deeper intermediate classifiers. A similar effect can also be achieved by adjusting the loss coefficients for deeper ICs, as proposed in prior work \citep{kaya2019shallow,han2022learning}, or by scaling the gradients of each IC, as proposed by Li et al. \citep{li2019improved}. We revisit these techniques within both the joint and mixed regimes to evaluate their impact, with the results presented in~\Cref{tab:loss_and_gradient_scaling}.

We first evaluate the loss scaling method proposed by Kaya et al.\citep{kaya2019shallow} (SDN) by training multi-exit ResNet-50 models on the TinyImageNet dataset. We also test \citep{han2022learning} constant loss weighting scheme, which adjusts the loss coefficients by either increasing (Inc.) or decreasing (Dec.) them along the model's depth. These schemes allow the model to prioritize different computational budgets. However, \textbf{the improvements observed in the mixed regime are notably smaller compared to those seen in the joint regime.}

Finally, we evaluate the gradient equilibrium method \citep{li2019improved}. For the joint regime, gradient scaling improves performance at higher computational budgets. However, for the mixed regime, there is no additional benefit, further confirming that they achieve a similar effect. \textbf{The mixed regime achieve superior results and it obviates the need for application of gradient equilibrium.}

\begin{table}
    \centering
    \caption{Accuracy improvement on the Tinyimagenet dataset when using various IC loss and gradient scaling methods. While gradient equilibrium enhances performance in the joint regime, it does not benefit the mixed regime, highlighting the effectiveness of the mixed regime for training multi-exit models.}
    \label{tab:loss_and_gradient_scaling}
    \begin{tabular}{cccccc}
        % BARE RESULTS version:
        % \toprule
        % Regime & Scaling & 25\% & 50\% & 75\% & 100\% \\
        % \midrule
        % \multirow{5}{*}{Joint} & - & $52.98$ & $62.16$ & $65.14$ & $65.01$ \\
        % & GE & $53.29$ & $62.56$ & $65.87$ & $65.72$ \\
        % & Inc. & $51.28$ & $62.56$ & $66.12$ & $66.07$ \\
        % & Dec. & $53.58$ & $61.87$ & $65.06$ & $65.01$ \\
        % & SDN & $49.23$ & $61.83$ & $66.03$ & $65.95$ \\
        % \midrule
        % \multirow{5}{*}{Mixed} & - & $52.89$ & $63.28$ & $67.20$ & $67.24$ \\
        % & GE & $51.73$ & $63.17$ & $67.22$ & $67.22$ \\
        % & Inc. & $51.81$ & $63.01$ & $67.42$ & $67.49$ \\
        % & Dec. & $53.58$ & $63.60$ & $66.76$ & $66.78$ \\
        % & SDN & $49.50$ & $62.18$ & $67.27$ & $67.47$ \\
        % \bottomrule
        % DIFF version:
        % \toprule
        % Regime & Scaling & 25\% & 50\% & 75\% & 100\% \\
        % \midrule
        % \multirow{5}{*}{Joint} & - & $52.98$ & $62.16$ & $65.14$ & $65.01$ \\
        % & GE & $+0.31$ & $+0.4$ & $+0.73$ & $+0.71$ \\
        % & Inc. & $-1.7$ & $+0.4$ & $+0.98$ & $+1.06$ \\
        % & Dec. & $+0.6$ & $-0.29$ & $-0.08$ & $+0.0$ \\
        % & SDN & $-3.75$ & $-0.33$ & $+0.89$ & $+0.94$ \\
        % \midrule
        % \multirow{5}{*}{Mixed} & - & $52.89$ & $63.28$ & $67.20$ & $67.24$ \\
        % & GE & $-1.16$ & $-0.11$ & $+0.02$ & $-0.02$ \\
        % & Inc. & $-1.08$ & $-0.27$ & $+0.22$ & $+0.25$ \\
        % & Dec. & $+0.69$ & $+0.32$ & $-0.44$ & $-0.46$ \\
        % & SDN & $-3.39$ & $-1.1$ & $+0.07$ & $+0.23$ \\
        % \bottomrule
        % DIFF alternative grouping version:
        \toprule
        Regime & Scaling & 25\% & 50\% & 75\% & 100\% \\
        % \midrule
        % Joint & - & $52.98$ & $62.16$ & $65.14$ & $65.01$ \\
        % Mixed & - & $52.89$ & $63.28$ & $67.20$ & $67.24$ \\
        \midrule
        Joint & Inc. & $-1.7$ & $+0.4$ & $+0.98$ & $+1.06$ \\
        Mixed & Inc. & $-1.08$ & $-0.27$ & $+0.22$ & $+0.25$ \\
        \midrule
        Joint & Dec. & $+0.6$ & $-0.29$ & $-0.08$ & $+0.0$ \\
        Mixed & Dec. & $+0.69$ & $+0.32$ & $-0.44$ & $-0.46$ \\
        \midrule
        Joint & SDN & $-3.75$ & $-0.33$ & $+0.89$ & $+0.94$ \\
        Mixed & SDN & $-3.39$ & $-1.1$ & $+0.07$ & $+0.23$ \\
        \midrule
        Joint & GE & $+0.31$ & $+0.4$ & $+0.73$ & $+0.71$ \\
        Mixed & GE & $-1.16$ & $-0.11$ & $+0.02$ & $-0.02$ \\
        \bottomrule
    \end{tabular}
\end{table}

\subsection{Impact of IC Size}
\label{sec:ic_size}

The size of an internal classifier (IC) in early exit architectures refers to the number of layers and neurons within the internal classifier. Smaller ICs are computationally efficient and enable faster early exits with minimal overhead. Conversely, larger ICs offer greater capacity, potentially enhancing accuracy, but may offset the computational benefits of multi-exit models.

A recent study by \cite{wojcik2023zero} investigated the impact of IC size and found that larger heads significantly improve performance. However, this analysis focused solely on the disjoint training regime. In our study, we examine the effect of varying IC sizes using a ViT-T model trained on the TinyImagenet dataset. Each IC consists of either one or two fully connected layers with hidden dimensions of 1024 or 2048, followed by a softmax layer. As summarized in Table \ref{table:heads-sizes}, under the disjoint regime larger ICs indeed yield clear gains: moving from a single-layer IC with 1024 units to a two-layer IC with 2048 units can increase top-1 accuracy by up to 7.6 percentage points, and this effect is especially visible for lower budgets. However, in the mixed regime this trend reverses, and the change can result in an actual performance degradation. These results demonstrate that optimal IC architecture depends critically on the chosen training strategy, and that conclusions drawn under a disjoint setup may not transfer when the backbone is also trained.

\begin{table}
    \caption{The effect of varying head size on the ViT-T model. Larger heads translate to inferior performance for the mixed regime. In contrast, the disjoint model clearly benefits from larger ICs when small budgets are considered.}
    \label{table:heads-sizes}
    \centering
    \begin{tabular}{cccccc}
        \toprule
        Regime & IC arch. & 25\% & 50\% & 75\% & 100\% \\
        \midrule
        \multirow{3}{*}{Disjoint} & 1L & $26.35$ & $41.69$ & $53.06$ & $56.72$ \\
        & 2L-1024 & $33.85$ & \underline{$47.21$} & \underline{$54.97$} & $56.72$ \\
        & 2L-2048 & \underline{$33.96$} & $45.98$ & $54.74$ & $56.72$ \\
        \midrule
        \multirow{3}{*}{Joint} & 1L & $42.47$ & $53.24$ & $56.02$ & $56.03$ \\
        & 2L-1024 & \underline{$45.59$} & \underline{$55.11$} & \underline{$57.73$} & \underline{$57.66$} \\
        & 2L-2048 & $43.60$ & $53.89$ & $56.94$ & $57.00$ \\
        \midrule
        \multirow{3}{*}{Mixed} & 1L & $44.03$ & $\textbf{57.91}$ & $\textbf{60.42}$ & $\textbf{60.28}$ \\
        & 2L-1024 & $\textbf{44.92}$ & $57.11$ & $59.32$ & $59.22$ \\
        & 2L-2048 & $44.61$ & $56.94$ & $60.18$ & $60.15$ \\
        \bottomrule
    \end{tabular}
\end{table}

\subsection{Impact of IC Placement Scheme}

The scheme of placing the internal classifiers in early exit architectures refers to where these classifiers are inserted at different layers within the model. This can range from being placed at every layer to being placed at strategically selected layers. To explore the impact of the placement scheme on the final performance, we train a ResNet-50 model with different regimes for every placement scheme. An IC can be placed at any block from index 0 to 14 for this architecture; however, we do not place it at index 0, as this would entirely bypass the backbone. In the Every-$n$ placement scheme, an IC is inserted at every $n$-th block. In the Dense-Sparse configuration, ICs are placed at blocks: $[1, 2, 3, 4, 5, 6, 7, 11]$, while in the Sparse-Dense scheme, they are placed at blocks: $[1, 4, 8, 9, 10, 11, 12, 13, 14]$.

The placement scheme plays a significant role in shaping network performance across different training regimes, as shown in~\Cref{tab:placement_schemes}. While densely placed ICs generally ensure strong performance across a range of computational budgets, the model can be tailored to prioritize a particular budget by using the Dense-Sparse or Sparse-Dense schemes. In most cases, models trained with the mixed regime outperform those trained under other regimes. 

\begin{table}[ht!]
    \caption{The effect of varying head placement schemes on the SDN early-exit architecture with ResNet-50 as a backbone, trained on the TinyImagenet dataset.}
    \label{tab:placement_schemes}
    \centering
    \begin{tabular}{lllllll}
        \toprule
        Scheme & Regime & 25\% & 50\% & 75\% & 100\%\\
        \midrule
        \multirow{3}{*}{Every-$1$} & Disjoint & $38.92$ & $49.25$ & $60.10$ & $65.76$ \\
        & Joint & $52.20$ & $62.49$ & $65.52$ & $65.59$ \\
        & Mixed & \underline{$52.22$} & \underline{$63.03$} & \underline{$67.21$} & \underline{$67.35$} \\
        \midrule
        \multirow{3}{*}{Every-$2$} & Disjoint & $37.34$ & $48.03$ & $60.34$ & $65.65$ \\
        & Joint & $51.81$ & $62.60$ & $65.55$ & $65.38$ \\
        & Mixed & \underline{$52.41$} & \underline{$63.19$} & \underline{$67.14$} & \underline{$67.33$} \\
        \midrule
        \multirow{3}{*}{Every-$3$} & Disjoint & $-$ & $47.91$ & $60.95$ & $65.77$ \\
        & Joint & $-$ & $\textbf{63.33}$ & \underline{$67.22$} & \underline{$67.21$} \\
        & Mixed & $-$ & $62.52$ & $66.72$ & $66.71$ \\
        \midrule
        \multirow{3}{*}{Every-$4$} & Disjoint & $-$ & $41.32$ & $57.78$ & $65.78$ \\
        & Joint & $-$ & \underline{$62.54$} & $66.30$ & $66.27$ \\
        & Mixed & $-$ & $62.07$ & \underline{$67.08$} & \underline{$67.14$} \\
        \midrule
        \multirow{3}{*}{Every-$5$} & Disjoint & $-$ & $39.85$ & $56.20$ & $65.72$ \\
        & Joint & $-$ & $61.10$ & $65.61$ & $65.79$ \\
        & Mixed & $-$ & \underline{$61.95$} & $\textbf{67.33}$ & \underline{$67.40$} \\
        \midrule
        \multirow{3}{*}{Den.-Spa.} & Disjoint & $38.47$ & $50.64$ & $62.04$ & $65.74$ \\
        & Joint & $53.14$ & $62.23$ & $64.76$ & $64.93$ \\
        & Mixed & $\textbf{53.48}$ & \underline{$63.17$} & \underline{$66.24$} & \underline{$66.27$} \\
        \midrule
        \multirow{3}{*}{Spa.-Den.} & Disjoint & $37.12$ & $47.03$ & $59.79$ & $65.68$ \\
        & Joint & $50.47$ & $61.19$ & $65.36$ & $65.42$ \\
        & Mixed & \underline{$51.19$} & \underline{$62.27$} & \underline{$67.26$} & $\textbf{67.47}$ \\
        \bottomrule
    \end{tabular}
    \label{tab:density}
\end{table}

\subsection{Training time}
The ultimate aim of a majority of multi-exit models is achieving the Pareto frontier in model performance and computational cost of the model during inference~\citep{scardapane2020should,han2021dynamic}. As such, to prevent under-training, we apply an early-stopping criterion in every experiment in this work. However, this raises an interesting question: how long does it take for the training in each regime to converge?

We explore the training time of the ResNet-50 model trained on the TinyImagenet dataset (see \Cref{tab:main_results}). Disjoint training was the fastest and took $523$ ($\pm 156$, averaged over multiple seeds) epochs on average. Joint training converged after $1610$ ($\pm 395$), while mixed training required $1166$ ($\pm 136$) epochs. These results indicate that while disjoint training converges most rapidly, the mixed regime offers a substantial speed-up over joint training by alleviating the interference effect between intermediate classifiers.

\section{Related Work}

% intro paragraph
Early exiting is a notable application of the conditional computation paradigm \citep{bengio2013estimating}. While conceptually similar to earlier classifier cascades \citep{xu2014classifier,wang2017idk}, it differs in that all classifiers are integrated within a single model, enabling end-to-end training. The first multi-exit model was introduced by \citep{teerapittayanon2016branchynet}, and the field has expanded considerably since its inception.

% joint
Joint training is the most widely used and well-established strategy for early-exit models \citep{matsubara2022split}. This approach has been successfully applied to dynamic inference under various constraints, such as energy or time limitations \citep{wang2020dual}, and extended to diverse early-exit applications, including low-resolution classification \citep{xing2020early}, quality enhancement \citep{yang2020resolution}, and Question-Answering systems \citep{soldaini2020cascade}. While joint training has proven effective, several studies have demonstrated significant improvements through modifications of the training process. For instance, knowledge distillation from the final classifier to earlier internal classifiers has been shown to enhance their performance \citep{phuong2019distillation,li2019improved,liu2020fastbert}. Similarly, ensembling multiple intermediate classifiers can improve the prediction accuracy \citep{qendro2021early,sarti2023anticipate}. The Global Past-Future (GPF) method \citep{liao2021global} goes a step further and incorporates information from both earlier predictions and surrogate later predictions to improve inference. Additionally, recent works \citep{han2022learning,yu2023boosted,chataoui2023jointly} identify a train-test mismatch in conventional multi-exit approaches and propose strategies that address this issue, further enhancing the robustness of early-exit models.

% disjoint
SDN \cite{kaya2019shallow} was one of the first to explore the training of early-exit models through the pre-training of the architecture's backbone followed by separate training of the classifiers. Multiple subsequent works have focused on optimizing early-exit models based only on this setup \cite{wojcik2023zero,lahiany2022pteenet, liu2020fastbert}, potentially limiting the general applicability of their findings. For instance,~\cite{wolczyk2021zero} employ an ensembling technique that combines predictions from earlier internal classifiers, weights of which are trained in a separate, third training phase. \cite{lahiany2022pteenet} propose PTEENet, which augments pre-trained networks with confidence heads that dynamically adjust based on available resources and unlabeled data. 

% regimes and weighting losses and gradients
\citep{kaya2019shallow} were the first to explore both joint and disjoint training approaches for early-exit models. These approaches are also briefly reviewed in surveys such as \citep{scardapane2020should,matsubara2022split}. Furthermore, techniques like weighting the losses at each exit head \citep{zhou2020bert,kaya2019shallow,han2022learning} or scaling the individual gradients \citep{li2019improved} can be regarded as variations of the joint training paradigm. In the context of LLMs, a concurrent work of \citet{bae2024relaxed} includes an empirical comparison of joint and disjoint training strategies, with similar conclusions about the weaknesses of both approaches. To the best of our knowledge, our work is the first to present a broad and systematic analysis of the early-exit training strategies.

\section{Conclusion}
\label{sec:conclusion}

This study emphasizes the critical importance of training regimes for early-exit models, an aspect that has been largely neglected in prior research. Addressing this gap, we provide a detailed analysis and evaluation of various training approaches for early-exit architectures in deep neural networks. Our findings reveal that the manner in which the backbone and internal classifiers are trained has a significant impact on the performance and efficiency of these models. Furthermore, we show that when using the mixed training regime, gradient scaling is unnecessary. Below we summarize some practical takeaways for the selection of the training regime.

\textit{Mixed}. Mixed regime demonstrates substantial robustness across various factors, including different data modalities and early-exit approaches with varying exit criteria. Therefore, the mixed regime is generally preferred, combining the benefits of both disjoint and joint training. The mixed regime ensures that the backbone network is well-optimized before integrating internal classifiers, leading to improved computational efficiency and accuracy. It is particularly recommended for cases where the performance on medium and high computational budgets is the main requirement. 

\textit{Joint.} The joint regime is simple to implement and can perform well for small computational budgets. However, it generally underperforms, especially for medium to higher budgets, which are more relevant in practical applications.

\textit{Disjoint.} This regime is generally inferior compared to the others in most setups, but performs well for the highest budgets. It may be preferred when the backbone is shared, or the lack of resources prevents us from training the backbone network.

\section*{Acknowlegdements}
%TODO Piotr
The research was funded by the program Excellence Initiative –
Research University at the Jagiellonian University in Kraków.

This research has been supported by a grant from the Faculty of Mathematics and Computer Science under the Strategic Programme Excellence Initiative at Jagiellonian University.

This research was partially funded by National Science Centre, Poland, grant no 2022/45/B/ST6/02817.

We gratefully acknowledge Polish high-performance computing infrastructure PLGrid (HPC Center: ACK Cyfronet AGH) for providing computer facilities and support within computational grants no. PLG/2023/016270, PLG/2024/017185, and PLG/2024/017385, and the LUMI consortium through PLL/2024/07/017641.

The contribution of Bartosz Wójcik to this research was conducted at the Faculty of Mathematics and Computer Science, and the Doctoral School of Exact and Natural Sciences of the Jagiellonian University.

\section*{Impact Statement}

This paper presents work whose goal is to advance the field of Machine Learning. There are many potential societal consequences of our work, none of which we feel must be specifically highlighted here.

\bibliography{main}
\bibliographystyle{icml2025}

%%%%%%%%%%%%%%%%%%%%%%%%%%%%%%%%%%%%%%%%%%%%%%%%%%%%%%%%%%%%%%%%%%%%%%%%%%%%%%%
%%%%%%%%%%%%%%%%%%%%%%%%%%%%%%%%%%%%%%%%%%%%%%%%%%%%%%%%%%%%%%%%%%%%%%%%%%%%%%%
% APPENDIX
%%%%%%%%%%%%%%%%%%%%%%%%%%%%%%%%%%%%%%%%%%%%%%%%%%%%%%%%%%%%%%%%%%%%%%%%%%%%%%%
%%%%%%%%%%%%%%%%%%%%%%%%%%%%%%%%%%%%%%%%%%%%%%%%%%%%%%%%%%%%%%%%%%%%%%%%%%%%%%%
\newpage
\appendix
\onecolumn

\section{Alternative Regimes}
\label{sec:alt_regimes}

In addition to the training regimes discussed in the main paper, early-exit surveys describe alternative training strategies \cite{scardapane2020should, rahmath2024early}. Moreover, \cite{xin2021berxit} have proposed the "Alternating" approach, which also attempts to combine the advantages of joint and disjoint regimes.  Additionally, we propose a refinement of the mixed regime, termed the mixed-gradual training regime.

We describe these alternative regimes below:
\begin{itemize}[itemsep=1pt]
    \item \noindent\textbf{Branch-wise training.} In this regime, ICs are trained sequentially, with only the part of the backbone corresponding to the current IC being unfrozen.
    \item \noindent\textbf{Separate training.} Similar to branch-wise training, ICs are optimized sequentially. However, at each step, the preceding ICs are jointly trained along with the current one. The entire model is unfrozen during the training process.
    \item \noindent\textbf{Alternating training.} Proposed in \cite{xin2021berxit}, this regime alternates between training the backbone independently (\Cref{eq:ph1}) and jointly training the backbone with all ICs (\Cref{eq:ph2}) in each training step.
    \item \noindent\textbf{Mixed-gradual training.} Building on the strengths of the mixed regime, we propose a more gradual optimization approach. Training proceeds in $m$ phases, where $m$ is the number of ICs. In the $i$-th phase, the last $i$ ICs are jointly optimized, and no weights are frozen at any stage. This ensures flexibility throughout the training process while allowing the model to adapt more effectively to the optimization landscape. This approach is similar to \emph{gradual early exit curriculum} proposed by the concurrent work of \citet{elhoushi2024layerskip}, with the difference being that we utilize early-stopping, while they enable the next IC at constant intervals of training steps.
\end{itemize}

In~\Cref{sec:full_results}, we present the results for these alternative training regimes. Both the branch-wise and separate training approaches demonstrate subpar performance when evaluated under higher computational budgets. Although the alternating training regime resembles the mixed regime in its motivation, its performance consistently falls short of both the mixed and joint regimes. Interestingly, the mixed-gradual regime outperforms the mixed regime, providing strong evidence in favor of our approach to enhancing early-exit optimization.

\section{Loss Landscape}

The concept of a loss landscape in the context of neural networks is crucial for understanding the training dynamics and generalization properties of models. The loss landscape provides a visual and analytical representation of how the loss function changes with respect to the model's parameters. By visualizing the loss landscapes of different neural network architectures, we can understand how design choices affect the shape of the loss function.

For a trained model with parameters $\theta^*$, one can evaluate the loss function for the numbers $x, y$
\begin{equation}
    f(x, y) = L(\theta^* + x\delta + y\eta)
\label{eq:loss_landscape}
\end{equation}
such that $\delta,\eta$ are random directions sampled from a probability distribution, usually a Gaussian distribution, filter-normalized \cite{li2018visualizing}, obtaining a 3D plot.
In contrast to a typical neural network architecture, in early-exit set-up, both the final and internal classifiers are considered. We consider total training loss and separate losses for each IC. 

When evaluating head losses, we use common random directions($\delta, \eta$) for each IC. The $\delta,\eta$ directions contain both backbone and head parameters.

\begin{figure}[h!]
    \centering
    % First row
    \begin{subfigure}{0.32\textwidth}
        \includegraphics[width=\linewidth, trim=80 20 100 50, clip]{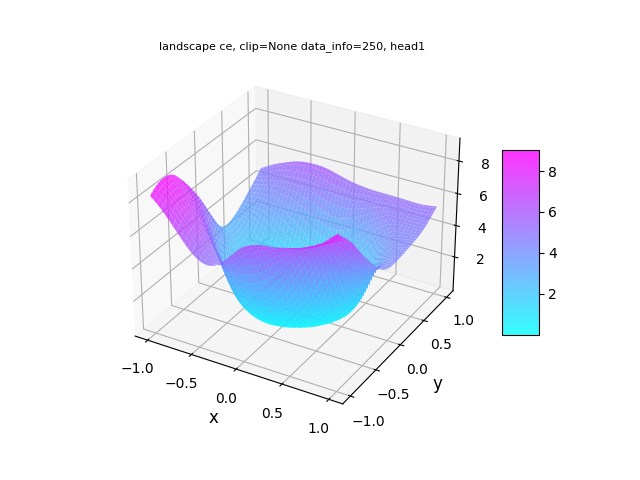}
    \end{subfigure}
    \hfill
    \begin{subfigure}{0.32\textwidth}
        \includegraphics[width=\linewidth, trim=80 20 100 50, clip]{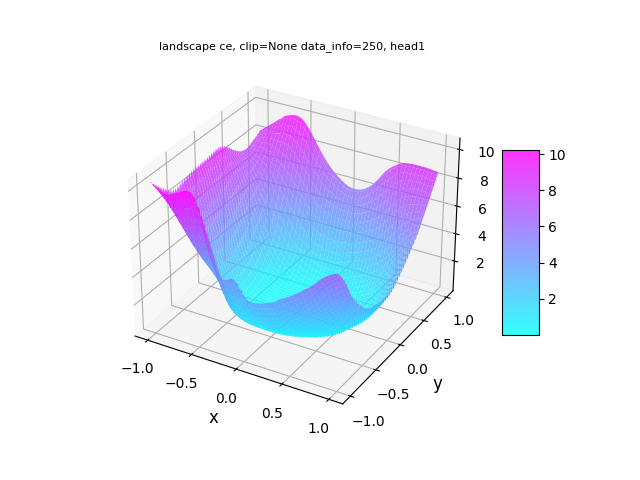}
    \end{subfigure}
    \hfill
    \begin{subfigure}{0.32\textwidth}
        \includegraphics[width=\linewidth, trim=80 20 100 50, clip]{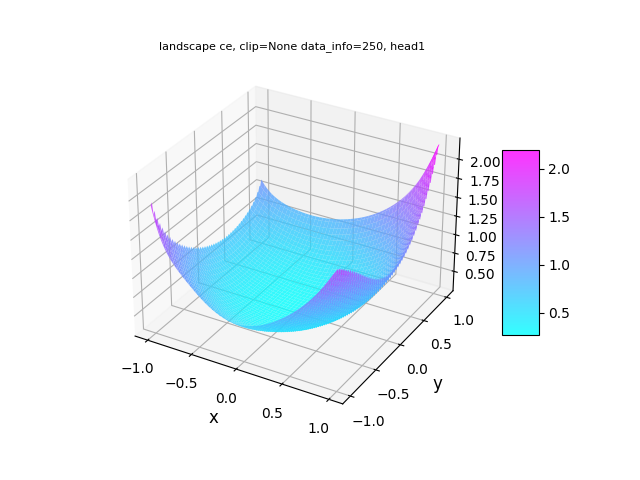}
    \end{subfigure}
    \caption{Training loss landscapes: comparison for Joint, Mixed, and Disjoint regimes (left to right), head 1. Landscapes for SDN architecture with Resnet20 as backbone trained on the CIFAR-10 dataset.}
    \label{fig:loss_landscape_regimes_comparison}
\end{figure}
As shown in ~\Cref{fig:loss_landscape_regimes_comparison} there is a significant difference in loss landscapes between the Disjoint regime and the Joint one. The Joint and Mixed regimes are similar in this regard.

\section{Full Results}
\label{sec:full_results}

Due to space constraints in the main paper we have presented only the averaged results. In this section, we present the main results with the standard deviation reported for each experiment. Furthermore, we include the results for additional training regimes, which are described in~\Cref{sec:alt_regimes}.

\begin{table}[H]
    \centering
    \caption{ResNet-34 CIFAR-100}
    \begin{tabular}{cccccc}
        \toprule
        Regime & 25\% & 50\% & 75\% & 100\% & Unlimited \\
        \midrule
        Disjoint & $52.57$ $ \pm 0.07$ & $67.56$ $ \pm 0.22$ & $73.49$ $ \pm 0.93$ & $73.79$ $ \pm 1.00$ & $73.79$ $ \pm 1.00$ \\
        Joint & $62.84$ $ \pm 0.67$ & $72.80$ $ \pm 0.34$ & $74.32$ $ \pm 0.08$ & $74.17$ $ \pm 0.10$ & $74.19$ $ \pm 0.09$ \\
        Mixed & $62.24$ $ \pm 0.78$ & $73.81$ $ \pm 0.51$ & $75.92$ $ \pm 0.44$ & $75.88$ $ \pm 0.40$ & $75.88$ $ \pm 0.40$ \\
        Branch-wise & $64.19$ $ \pm 0.47$ & $67.03$ $ \pm 0.76$ & $66.88$ $ \pm 0.60$ & $66.75$ $ \pm 0.68$ & $66.75$ $ \pm 0.68$ \\
        Separate & $66.25$ $ \pm 0.54$ & $72.57$ $ \pm 0.38$ & $72.87$ $ \pm 0.41$ & $72.82$ $ \pm 0.46$ & $72.82$ $ \pm 0.46$ \\
        Alternating & $60.35$ $ \pm 1.15$ & $71.99$ $ \pm 0.45$ & $74.57$ $ \pm 0.17$ & $74.37$ $ \pm 0.02$ & $74.36$ $ \pm 0.09$ \\
        Mixed-gradual & $60.11$ $ \pm 0.68$ & $73.16$ $ \pm 0.70$ & $76.08$ $ \pm 0.54$ & $75.94$ $ \pm 0.75$ & $75.94$ $ \pm 0.75$ \\
        \bottomrule
    \end{tabular}
    \label{tab:full_cifar100_resnet34}
\end{table}

\begin{table}[H]
    \centering
    \caption{MSDNet CIFAR-100}
    \begin{tabular}{cccccc}
        \toprule
        Regime & 25\% & 50\% & 75\% & 100\% & Unlimited \\
        \midrule
        Disjoint & $56.74$ $ \pm 0.41$ & $63.96$ $ \pm 0.74$ & $68.59$ $ \pm 0.72$ & $70.36$ $ \pm 0.84$ & $70.54$ $ \pm 0.93$ \\
        Joint & $65.93$ $ \pm 0.20$ & $72.02$ $ \pm 0.27$ & $74.73$ $ \pm 0.18$ & $75.86$ $ \pm 0.26$ & $75.93$ $ \pm 0.21$ \\
        Mixed & $65.94$ $ \pm 0.55$ & $72.01$ $ \pm 0.47$ & $75.46$ $ \pm 0.15$ & $76.51$ $ \pm 0.08$ & $76.71$ $ \pm 0.17$ \\
        Branch-wise & $58.24$ $ \pm 0.45$ & $59.73$ $ \pm 0.41$ & $60.16$ $ \pm 0.47$ & $59.97$ $ \pm 0.38$ & $59.73$ $ \pm 0.29$ \\
        Separate & $64.90$ $ \pm 0.56$ & $67.84$ $ \pm 0.17$ & $69.02$ $ \pm 0.25$ & $69.21$ $ \pm 0.23$ & $69.15$ $ \pm 0.23$ \\
        Alternating & $64.69$ $ \pm 0.45$ & $71.67$ $ \pm 0.27$ & $74.93$ $ \pm 0.16$ & $76.06$ $ \pm 0.25$ & $76.34$ $ \pm 0.31$ \\
        Mixed-gradual & $64.73$ $ \pm 0.66$ & $72.10$ $ \pm 0.47$ & $75.66$ $ \pm 0.51$ & $77.21$ $ \pm 0.61$ & $77.49$ $ \pm 0.64$ \\
        \bottomrule
    \end{tabular}
    \label{tab:full_cifar100_msdnet}
\end{table}

\begin{table}[H]
    \centering
    \caption{ViT-T CIFAR-100}
    \begin{tabular}{cccccc}
        \toprule
        Regime & 25\% & 50\% & 75\% & 100\% & Unlimited \\
        \midrule
        Disjoint & $28.33$ $ \pm 0.47$ & $47.32$ $ \pm 0.81$ & $61.50$ $ \pm 1.82$ & $63.99$ $ \pm 1.68$ & $63.99$ $ \pm 1.68$ \\
        Joint & $40.87$ $ \pm 0.84$ & $60.48$ $ \pm 0.94$ & $66.22$ $ \pm 1.07$ & $66.49$ $ \pm 1.08$ & $66.49$ $ \pm 1.08$ \\
        Mixed & $41.65$ $ \pm 0.19$ & $64.07$ $ \pm 0.38$ & $70.09$ $ \pm 0.64$ & $70.25$ $ \pm 0.56$ & $70.25$ $ \pm 0.56$ \\
        Branch-wise & $32.41$ $ \pm 1.32$ & $33.46$ $ \pm 0.76$ & $34.33$ $ \pm 0.49$ & $34.51$ $ \pm 0.45$ & $34.51$ $ \pm 0.45$ \\
        Separate & $42.67$ $ \pm 0.89$ & $51.48$ $ \pm 3.58$ & $54.29$ $ \pm 4.68$ & $54.42$ $ \pm 4.83$ & $54.42$ $ \pm 4.83$ \\
        Alternating & $37.85$ $ \pm 0.40$ & $59.11$ $ \pm 0.20$ & $67.81$ $ \pm 0.69$ & $68.39$ $ \pm 0.78$ & $68.39$ $ \pm 0.78$ \\
        Mixed-gradual & $40.61$ $ \pm 0.08$ & $64.83$ $ \pm 0.46$ & $71.42$ $ \pm 0.34$ & $71.60$ $ \pm 0.38$ & $71.60$ $ \pm 0.38$ \\
        \bottomrule
    \end{tabular}
    \label{tab:full_cifar100_vit_t}
\end{table}

\begin{table}[H]
    \centering
    \caption{ResNet-50 Tinyimagenet}
    \begin{tabular}{cccccc}
        \toprule
        Regime & 35\% & 50\% & 75\% & 100\% & Unlimited \\
        \midrule
        Disjoint & $38.49$ $ \pm 0.46$ & $49.25$ $ \pm 1.33$ & $60.50$ $ \pm 1.27$ & $65.71$ $ \pm 0.90$ & $65.80$ $ \pm 0.90$ \\
        Joint & $52.98$ $ \pm 0.77$ & $62.16$ $ \pm 0.61$ & $65.14$ $ \pm 0.57$ & $65.01$ $ \pm 0.75$ & $65.00$ $ \pm 0.75$ \\
        Mixed & $52.89$ $ \pm 0.23$ & $63.28$ $ \pm 0.59$ & $67.20$ $ \pm 0.65$ & $67.24$ $ \pm 0.71$ & $67.22$ $ \pm 0.71$ \\
        \bottomrule
    \end{tabular}
    \label{tab:full_resnet50_tinyimagenet}
\end{table}

\begin{table}[H]
    \centering
    \caption{ViT-T ImageNet-1k}
    \begin{tabular}{cccccc}
        \toprule
        Regime & 25\% & 50\% & 75\% & 100\% & Unlimited \\
        \midrule
        Disjoint & $10.22$ $ \pm 0.15$ & $35.15$ $ \pm 0.97$ & $61.23$ $ \pm 1.37$ & $71.61$ $ \pm 0.68$ & $71.61$ $ \pm 0.68$ \\
        Joint & $36.39$ $ \pm 0.12$ & $61.89$ $ \pm 0.71$ & $68.08$ $ \pm 0.84$ & $68.39$ $ \pm 0.87$ & $68.39$ $ \pm 0.87$ \\
        Mixed & $35.91$ $ \pm 0.19$ & $62.96$ $ \pm 0.12$ & $70.79$ $ \pm 0.36$ & $71.20$ $ \pm 0.34$ & $71.20$ $ \pm 0.34$ \\
        \bottomrule
    \end{tabular}
    \label{tab:full_vit_t_imagenet1k}
\end{table}

\begin{table}[H]
    \centering
    \caption{BERT-B 20-Newsgroups}
    \begin{tabular}{cccccc}
        \toprule
        Regime & 25\% & 50\% & 75\% & 100\% & Unlimited \\
        \midrule
        Disjoint & $68.53$ $ \pm 0.97$ & $83.75$ $ \pm 0.34$ & $85.75$ $ \pm 0.16$ & $85.69$ $ \pm 0.28$ & $85.54$ $ \pm 0.34$ \\
        Joint & $84.24$ $ \pm 0.48$ & $84.41$ $ \pm 0.41$ & $84.41$ $ \pm 0.41$ & $84.41$ $ \pm 0.41$ & $84.51$ $ \pm 0.60$ \\
        Mixed & $84.99$ $ \pm 0.84$ & $85.25$ $ \pm 0.64$ & $85.25$ $ \pm 0.64$ & $85.25$ $ \pm 0.64$ & $85.46$ $ \pm 0.44$ \\
        Branch-wise & $80.20$ $ \pm 0.19$ & $80.05$ $ \pm 0.12$ & $80.05$ $ \pm 0.12$ & $80.05$ $ \pm 0.12$ & $79.83$ $ \pm 0.06$ \\
        Separate & $79.87$ $ \pm 0.36$ & $79.87$ $ \pm 0.36$ & $79.87$ $ \pm 0.36$ & $79.87$ $ \pm 0.36$ & $79.78$ $ \pm 0.30$ \\
        Alternating & $83.51$ $ \pm 0.33$ & $83.92$ $ \pm 0.06$ & $83.92$ $ \pm 0.06$ & $83.92$ $ \pm 0.06$ & $84.18$ $ \pm 0.34$ \\
        Mixed-gradual & $81.35$ $ \pm 1.24$ & $81.35$ $ \pm 1.24$ & $81.35$ $ \pm 1.24$ & $81.35$ $ \pm 1.24$ & $81.24$ $ \pm 0.93$ \\
        \bottomrule
    \end{tabular}
    \label{tab:full_bert_newsgroups}
\end{table}

\begin{table}[H]
    \centering
    \caption{BERT-B STS-B}
    \begin{tabular}{cccccc}
        \toprule
        Regime & 25\% & 50\% & 75\% & 100\% & Unlimited \\
        \midrule
        Disjoint & $2.37$ $ \pm 0.01$ & $1.55$ $ \pm 0.11$ & $0.54$ $ \pm 0.00$ & $0.51$ $ \pm 0.02$ & $0.51$ $ \pm 0.02$ \\
        Joint & $1.70$ $ \pm 0.23$ & $0.61$ $ \pm 0.02$ & $0.53$ $ \pm 0.01$ & $0.52$ $ \pm 0.02$ & $0.52$ $ \pm 0.01$ \\
        Mixed & $2.43$ $ \pm 0.03$ & $0.59$ $ \pm 0.01$ & $0.53$ $ \pm 0.01$ & $0.51$ $ \pm 0.00$ & $0.51$ $ \pm 0.00$ \\
        Branch-wise & $2.14$ $ \pm 0.24$ & $0.82$ $ \pm 0.09$ & $0.84$ $ \pm 0.10$ & $0.83$ $ \pm 0.10$ & $0.84$ $ \pm 0.10$ \\
        Separate & $2.69$ $ \pm 0.14$ & $1.41$ $ \pm 0.09$ & $1.41$ $ \pm 0.09$ & $1.40$ $ \pm 0.08$ & $1.39$ $ \pm 0.0$ \\
        Alternating & $1.76$ $ \pm 0.33$ & $0.60$ $ \pm 0.00$ & $0.51$ $ \pm 0.01$ & $0.50$ $ \pm 0.01$ & $0.50$ $ \pm 0.01$ \\
        Mixed-gradual & $2.37$ $ \pm 0.03$ & $0.51$ $ \pm 0.00$ & $0.48$ $ \pm 0.00$ & $0.46$ $ \pm 0.01$ & $0.46$ $ \pm 0.01$ \\
        \bottomrule
    \end{tabular}
    \label{tab:full_bert_stsb}
\end{table}

\begin{table}[H]
    \centering
    \caption{ViT-T PBEE CIFAR-100}
    \begin{tabular}{cccccc}
        \toprule
        Regime & 25\% & 50\% & 75\% & 100\% & Unlimited \\
        \midrule
        Disjoint & $19.28$ $ \pm 0.32$ & $44.11$ $ \pm 1.73$ & $57.85$ $ \pm 1.61$ & $63.91$ $ \pm 1.68$ & $63.99$ $ \pm 1.68$ \\
        Joint & $28.66$ $ \pm 0.50$ & $56.49$ $ \pm 1.48$ & $64.31$ $ \pm 0.87$ & $66.52$ $ \pm 1.08$ & $66.49$ $ \pm 1.08$ \\
        Mixed & $28.87$ $ \pm 0.44$ & $59.42$ $ \pm 0.50$ & $68.33$ $ \pm 0.60$ & $70.33$ $ \pm 0.68$ & $70.25$ $ \pm 0.56$ \\
        \bottomrule
    \end{tabular}
    \label{tab:full_pbee}
\end{table}

\begin{table}[H]
    \centering
    \caption{ViT-T GPF CIFAR-100}
    \begin{tabular}{cccccc}
        \toprule
        Regime & 25\% & 50\% & 75\% & 100\% & Unlimited \\
        \midrule
        Disjoint & $33.16$ $ \pm 0.64$ & $53.19$ $ \pm 0.66$ & $62.54$ $ \pm 1.59$ & $64.01$ $ \pm 1.67$ & $63.99$ $ \pm 1.68$ \\
        Joint & $46.21$ $ \pm 0.30$ & $62.23$ $ \pm 0.81$ & $66.84$ $ \pm 0.68$ & $67.19$ $ \pm 0.78$ & $67.19$ $ \pm 0.80$ \\
        Mixed & $47.21$ $ \pm 1.02$ & $64.51$ $ \pm 0.84$ & $68.92$ $ \pm 0.30$ & $69.10$ $ \pm 0.38$ & $69.10$ $ \pm 0.38$ \\
        \bottomrule
    \end{tabular}
    \label{tab:full_gpf}
\end{table}

\begin{table}[H]
    \centering
    \caption{ViT-T Entropy CIFAR-100}
    \begin{tabular}{cccccc}
        \toprule
        Regime & 25\% & 50\% & 75\% & 100\% & Unlimited \\
        \midrule
        Disjoint & $27.14$ $ \pm 0.36$ & $46.09$ $ \pm 0.59$ & $60.30$ $ \pm 1.41$ & $63.99$ $ \pm 1.68$ & $63.99$ $ \pm 1.68$ \\
        Joint & $41.16$ $ \pm 0.82$ & $58.60$ $ \pm 0.89$ & $65.69$ $ \pm 1.23$ & $66.49$ $ \pm 1.08$ & $66.49$ $ \pm 1.08$ \\
        Mixed & $41.54$ $ \pm 0.52$ & $62.20$ $ \pm 0.35$ & $69.63$ $ \pm 0.47$ & $70.25$ $ \pm 0.56$ & $70.25$ $ \pm 0.56$ \\
        \bottomrule
    \end{tabular}
    \label{tab:full_entropy}
\end{table}

\begin{table}[H]
    \centering
    \caption{ViT-B pretrained on ImageNet-1k to CIFAR-100}
    \begin{tabular}{cccccc}
        \toprule
        Regime & 35\% & 50\% & 75\% & 100\% & Unlimited \\
        \midrule
        Disjoint & $28.50$ $ \pm 2.02$ & $55.64$ $ \pm 3.71$ & $87.49$ $ \pm 0.40$ & $89.95$ $ \pm 0.18$ & $89.95$ $ \pm 0.18$ \\
        Joint & $73.18$ $ \pm 1.36$ & $85.41$ $ \pm 0.23$ & $87.82$ $ \pm 0.22$ & $87.85$ $ \pm 0.29$ & $87.84$ $ \pm 0.29$ \\
        Mixed & $73.38$ $ \pm 0.35$ & $85.70$ $ \pm 0.63$ & $88.06$ $ \pm 0.31$ & $88.23$ $ \pm 0.29$ & $88.23$ $ \pm 0.29$ \\
        \bottomrule
    \end{tabular}
    \label{tab:full_transfer_learning}
\end{table}

\begin{table}[H]
    \centering
    \caption{ResNet-50 Tinyimagenet loss and gradient scaling experiment}
    \begin{tabular}{cccccc}
        \toprule
        Regime & 25\% & 50\% & 75\% & 100\% & Unlimited \\
        \midrule
        Joint & $52.98$ $ \pm 0.77$ & $62.16$ $ \pm 0.61$ & $65.14$ $ \pm 0.57$ & $65.01$ $ \pm 0.75$ & $65.00$ $ \pm 0.75$ \\
        Joint GE  & $53.29$ $ \pm 0.90$ & $62.56$ $ \pm 0.15$ & $65.87$ $ \pm 0.36$ & $65.72$ $ \pm 0.56$ & $65.69$ $ \pm 0.57$ \\
        Joint Inc. & $51.28$ $ \pm 0.19$ & $62.56$ $ \pm 0.12$ & $66.12$ $ \pm 0.39$ & $66.07$ $ \pm 0.42$ & $66.06$ $ \pm 0.41$ \\
        Joint Dec. & $53.58$ $ \pm 0.44$ & $61.87$ $ \pm 0.45$ & $65.06$ $ \pm 0.35$ & $65.01$ $ \pm 0.29$ & $65.00$ $ \pm 0.30$ \\
        Joint SDN & $49.23$ $ \pm 0.64$ & $61.83$ $ \pm 0.32$ & $66.03$ $ \pm 0.18$ & $65.95$ $ \pm 0.25$ & $65.93$ $ \pm 0.24$ \\
        Mixed & $52.89$ $ \pm 0.23$ & $63.28$ $ \pm 0.59$ & $67.20$ $ \pm 0.65$ & $67.24$ $ \pm 0.71$ & $67.22$ $ \pm 0.71$ \\
        Mixed GE  & $51.73$ $ \pm 0.12$ & $63.17$ $ \pm 0.33$ & $67.22$ $ \pm 0.39$ & $67.22$ $ \pm 0.10$ & $67.21$ $ \pm 0.12$ \\
        Mixed Inc. & $51.81$ $ \pm 0.24$ & $63.01$ $ \pm 0.49$ & $67.42$ $ \pm 0.39$ & $67.49$ $ \pm 0.32$ & $67.47$ $ \pm 0.34$ \\
        Mixed Dec. & $53.58$ $ \pm 0.69$ & $63.60$ $ \pm 0.25$ & $66.76$ $ \pm 0.50$ & $66.78$ $ \pm 0.47$ & $66.75$ $ \pm 0.47$ \\
        Mixed SDN & $49.50$ $ \pm 0.74$ & $62.18$ $ \pm 0.76$ & $67.27$ $ \pm 0.68$ & $67.47$ $ \pm 0.55$ & $67.47$ $ \pm 0.56$ \\
        \bottomrule
    \end{tabular}
    \label{tab:full_gradient_loss_scaling}
\end{table}

\begin{table}[H]
    \centering
    \caption{ViT-T Imagenette head size experiment}
    % \caption{ViT-T Tinyimagenet head size experiment}
    \begin{tabular}{cccccc}
        \toprule
        Regime, Head & 25\% & 50\% & 75\% & 100\% & Unlimited \\
        \midrule
        % Disjoint 1L & $24.91$ $ \pm 0.97$ & $39.06$ $ \pm 0.33$ & $51.86$ $ \pm 0.75$ & $55.96$ $ \pm 0.79$ & $55.96$ $ \pm 0.79$ \\
        % Disjoint 2L-1024 & $29.91$ $ \pm 0.68$ & $43.82$ $ \pm 0.22$ & $53.20$ $ \pm 0.55$ & $55.95$ $ \pm 0.79$ & $55.96$ $ \pm 0.79$ \\
        % Disjoint 2L-2048 & $29.45$ $ \pm 0.55$ & $43.32$ $ \pm 0.81$ & $52.97$ $ \pm 0.60$ & $55.95$ $ \pm 0.80$ & $55.96$ $ \pm 0.79$ \\
        % Joint 1L & $42.51$ $ \pm 0.28$ & $54.11$ $ \pm 0.18$ & $56.84$ $ \pm 0.09$ & $56.68$ $ \pm 0.05$ & $56.68$ $ \pm 0.05$ \\
        % Joint 2L-1024 & $45.28$ $ \pm 0.44$ & $55.59$ $ \pm 0.29$ & $58.11$ $ \pm 0.18$ & $58.13$ $ \pm 0.31$ & $58.12$ $ \pm 0.29$ \\
        % Joint 2L-2024 & $44.50$ $ \pm 1.18$ & $56.35$ $ \pm 0.84$ & $58.85$ $ \pm 0.39$ & $58.98$ $ \pm 0.32$ & $58.97$ $ \pm 0.32$ \\
        % Mixed 1L & $44.85$ $ \pm 0.55$ & $58.13$ $ \pm 0.36$ & $59.76$ $ \pm 1.30$ & $59.19$ $ \pm 1.80$ & $59.19$ $ \pm 1.80$ \\
        % Mixed 2L-1024 & $45.82$ $ \pm 0.43$ & $56.98$ $ \pm 0.26$ & $59.04$ $ \pm 0.25$ & $58.80$ $ \pm 0.67$ & $58.80$ $ \pm 0.66$ \\
        % Mixed 2L-2048 & $45.54$ $ \pm 0.69$ & $56.88$ $ \pm 0.37$ & $58.90$ $ \pm 0.23$ & $58.68$ $ \pm 0.65$ & $58.68$ $ \pm 0.66$ \\

        % Imagenette
        Disjoint 1L & $75.61$ $ \pm 0.10$ & $78.72$ $ \pm 1.24$ & $78.32$ $ \pm 1.62$ & $78.29$ $ \pm 1.63$ & $78.28$ $ \pm 1.65$ \\
        Disjoint 2L-1024 & $77.41$ $ \pm 0.39$ & $79.04$ $ \pm 1.26$ & $78.36$ $ \pm 1.63$ & $78.29$ $ \pm 1.63$ & $78.28$ $ \pm 1.65$ \\
        Disjoint 2L-2048 & $77.10$ $ \pm 0.40$ & $79.02$ $ \pm 1.26$ & $78.39$ $ \pm 1.69$ & $78.30$ $ \pm 1.65$ & $78.28$ $ \pm 1.65$ \\
        Joint 1L & $81.44$ $ \pm 1.54$ & $82.51$ $ \pm 1.40$ & $82.46$ $ \pm 1.41$ & $82.46$ $ \pm 1.41$ & $82.45$ $ \pm 1.40$ \\
        Joint 2L-1024 & $80.00$ $ \pm 0.69$ & $81.06$ $ \pm 0.57$ & $80.99$ $ \pm 0.69$ & $80.99$ $ \pm 0.69$ & $81.00$ $ \pm 0.68$ \\
        Joint 2L-2048 & $79.68$ $ \pm 0.28$ & $80.67$ $ \pm 0.08$ & $80.64$ $ \pm 0.15$ & $80.64$ $ \pm 0.15$ & $80.61$ $ \pm 0.13$ \\
        Mixed 1L & $82.21$ $ \pm 0.08$ & $83.20$ $ \pm 0.39$ & $83.19$ $ \pm 0.41$ & $83.19$ $ \pm 0.41$ & $83.18$ $ \pm 0.45$ \\
        Mixed 2L-1024 & $80.25$ $ \pm 0.59$ & $81.28$ $ \pm 0.44$ & $81.13$ $ \pm 0.40$ & $81.13$ $ \pm 0.40$ & $81.11$ $ \pm 0.37$ \\
        Mixed 2L-2048 & $79.63$ $ \pm 0.67$ & $80.77$ $ \pm 0.58$ & $80.70$ $ \pm 0.51$ & $80.70$ $ \pm 0.51$ & $80.67$ $ \pm 0.52$ \\
        \bottomrule
    \end{tabular}
    \label{tab:full_vit_imagenette}
\end{table}

\section{Results as Performance-Cost plots}
Following \citet{kaya2019shallow}, we presented most of the results in this paper in the form of tables. In this section, we provide the same results as model performance vs. computational costs plots.

\begin{figure}[h!]
    \centering

    % 1st row
    \begin{minipage}{0.47\textwidth}
        \centering
        \includegraphics[width=\linewidth]{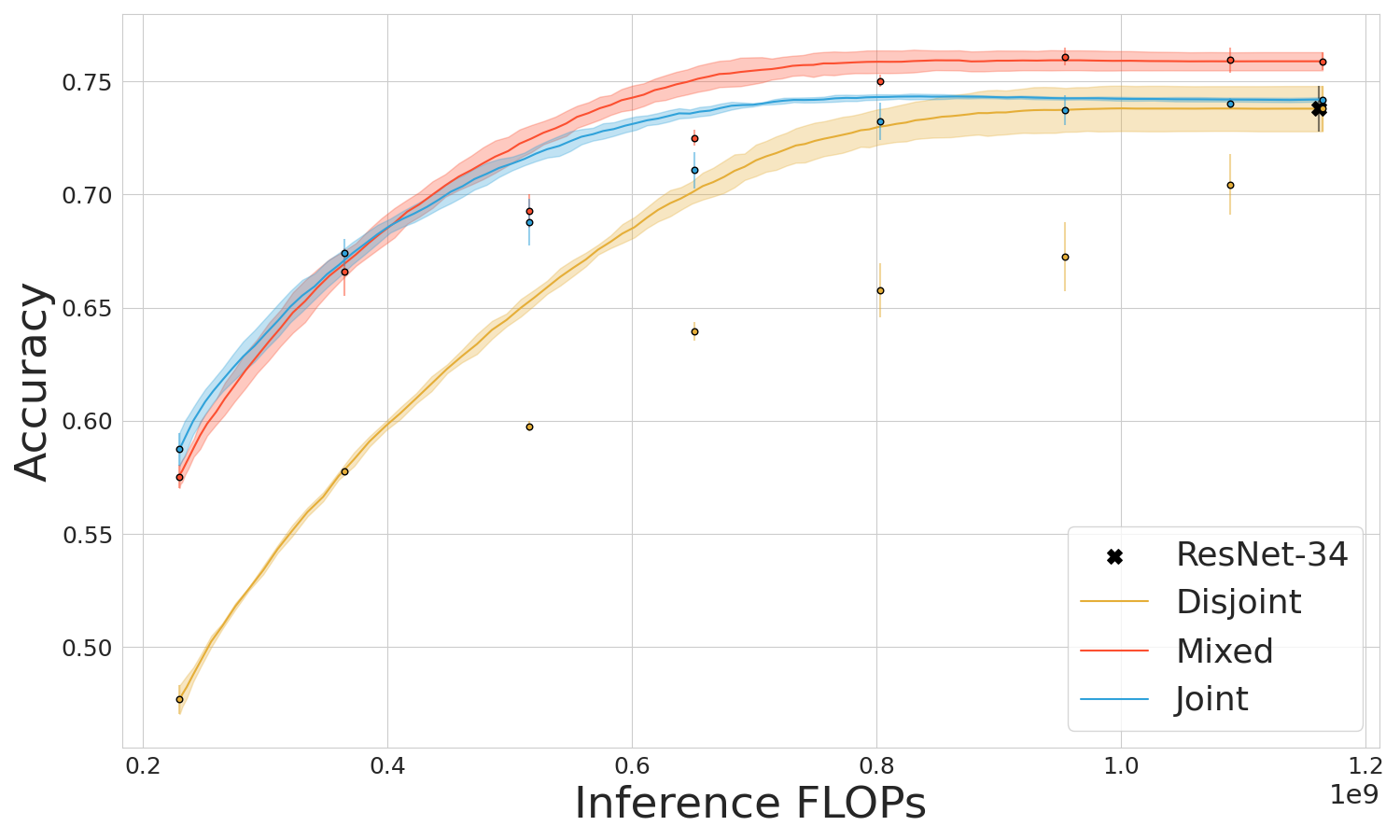}
        \caption{Performance–cost trade-off of the multi-exit ResNet-34 on the CIFAR-100 dataset}
    \end{minipage}
    \hfill
    \begin{minipage}{0.47\textwidth}
        \centering
        \includegraphics[width=\linewidth]{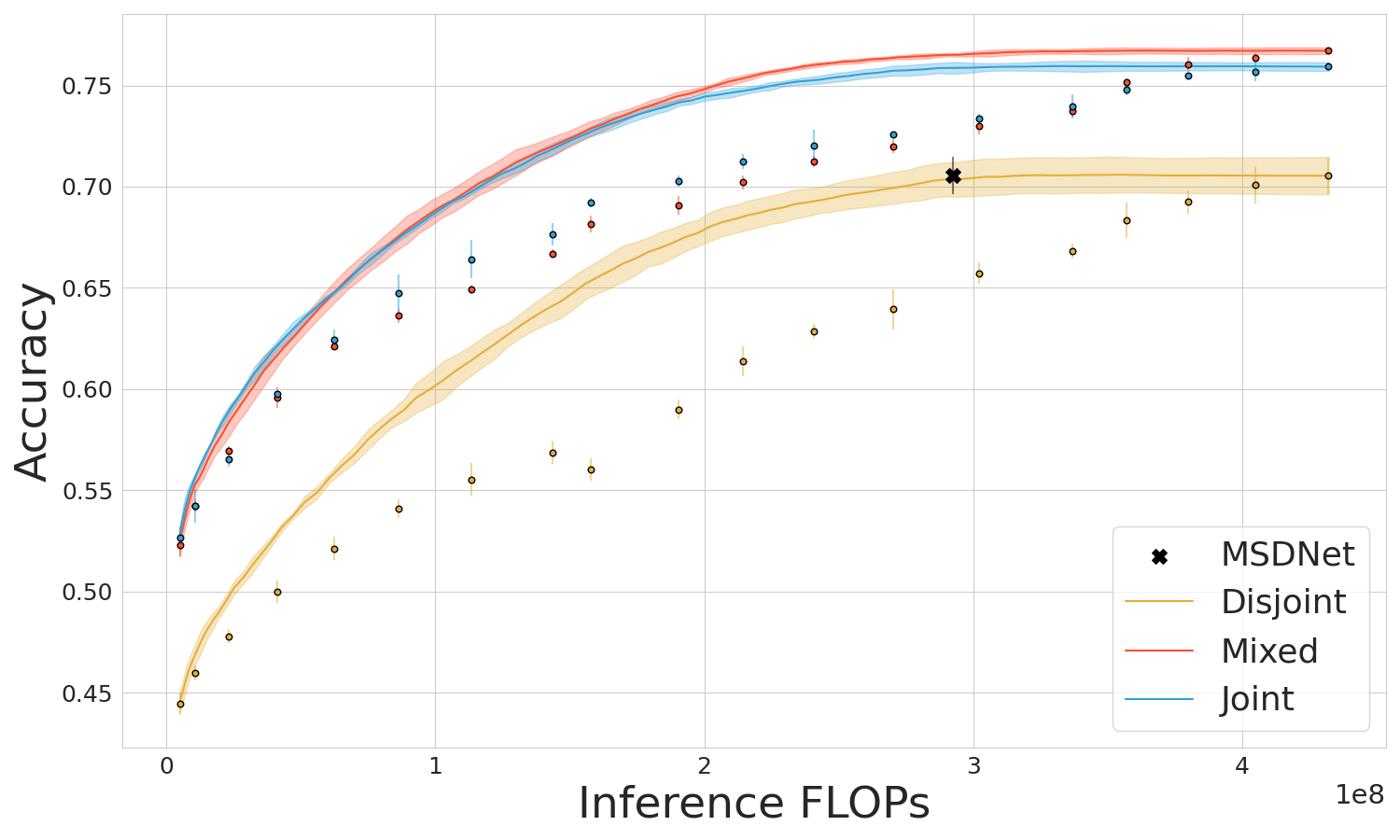}
        \caption{Performance–cost trade-off of the MSDNet architecture on the CIFAR-100 dataset}
    \end{minipage}

    \vspace{1.2em}

    % 2nd row
    \begin{minipage}{0.47\textwidth}
        \centering
        \includegraphics[width=\linewidth]{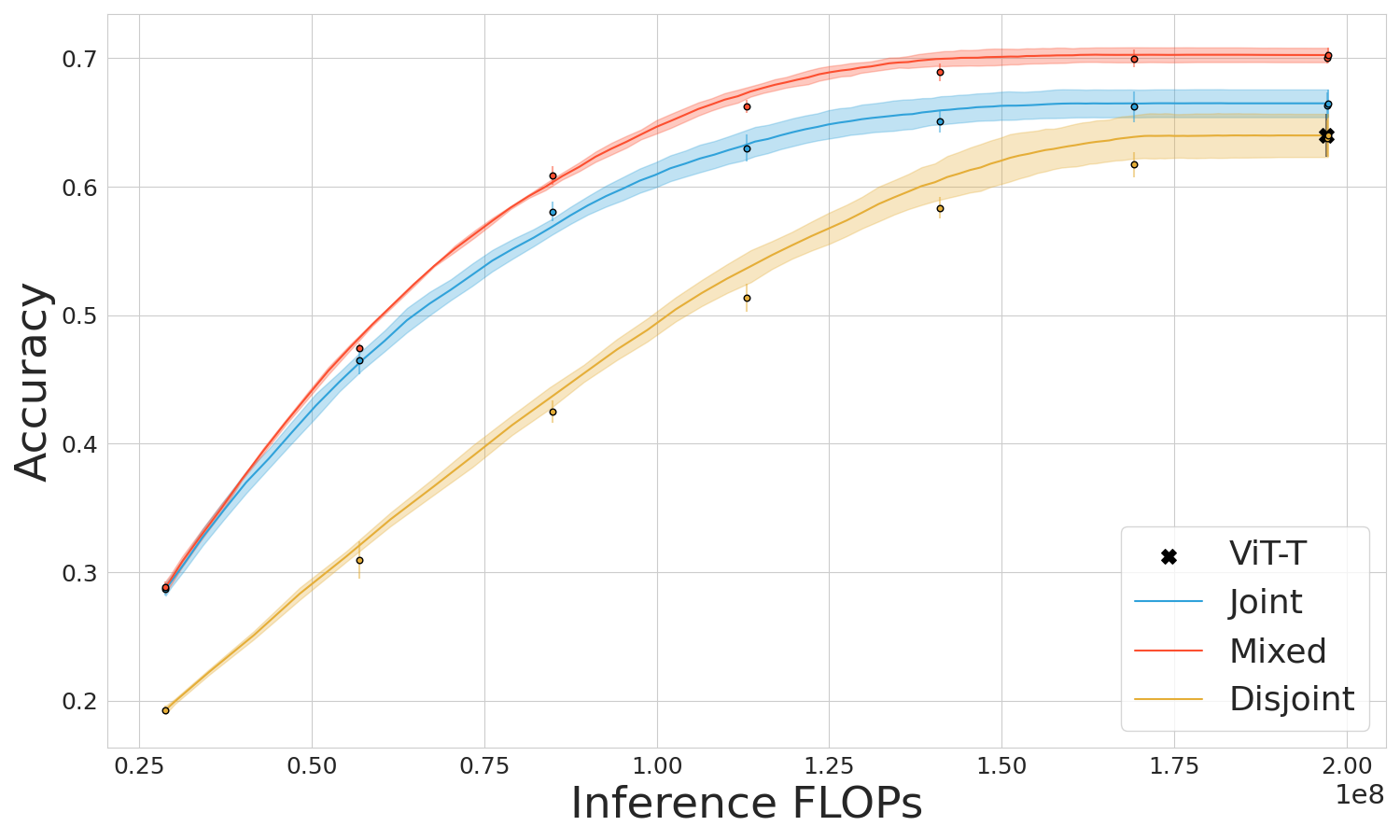}
        \caption{Performance–cost trade-off of the multi-exit ViT-T on the CIFAR-100 dataset}
    \end{minipage}
    \hfill
    \begin{minipage}{0.47\textwidth}
        \centering
        \includegraphics[width=\linewidth]{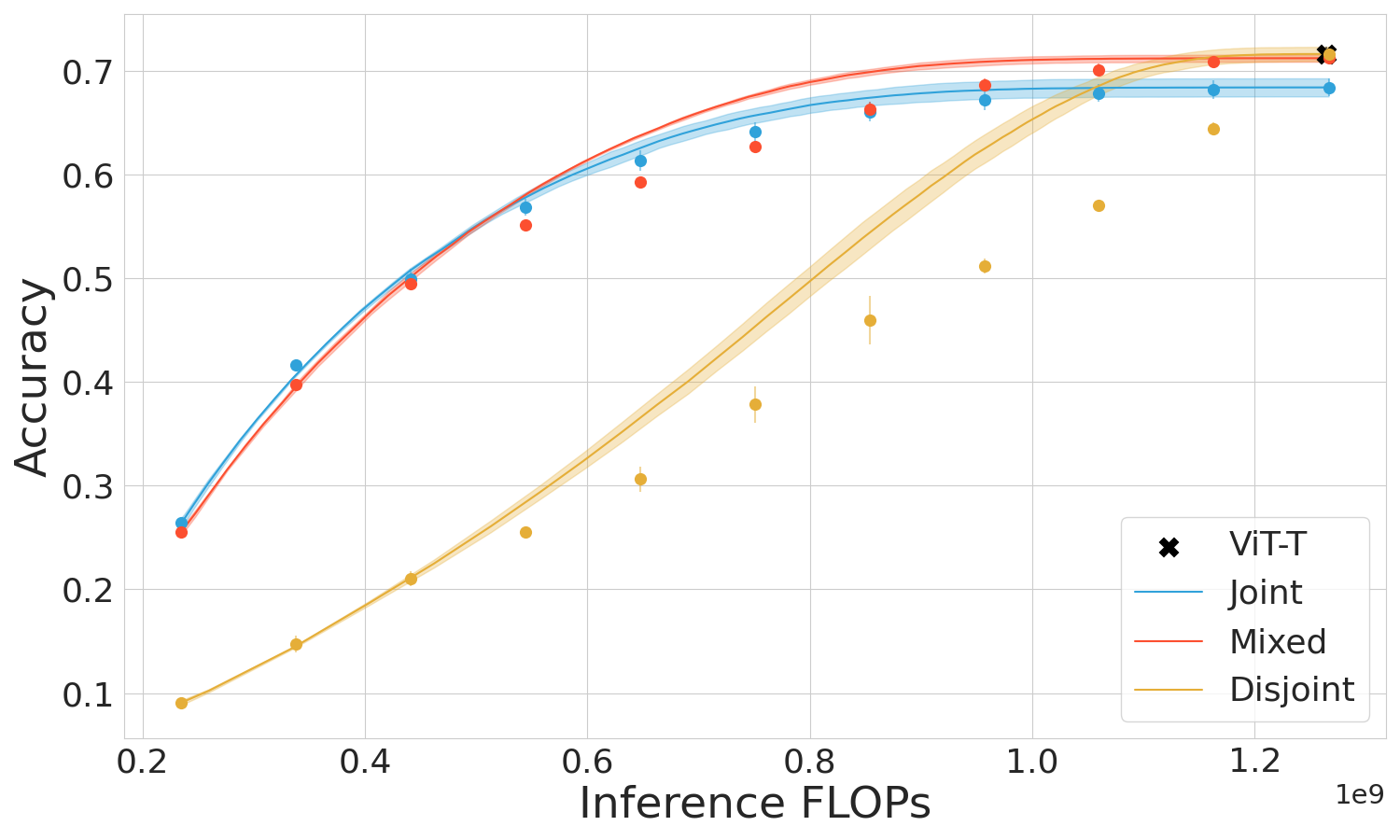}
        \caption{Performance–cost trade-off of the ViT-T architecture on the ImageNet-1k dataset}
    \end{minipage}

    \vspace{1.2em}

    % 3rd row
    \begin{minipage}{0.47\textwidth}
        \centering
        \includegraphics[width=\linewidth]{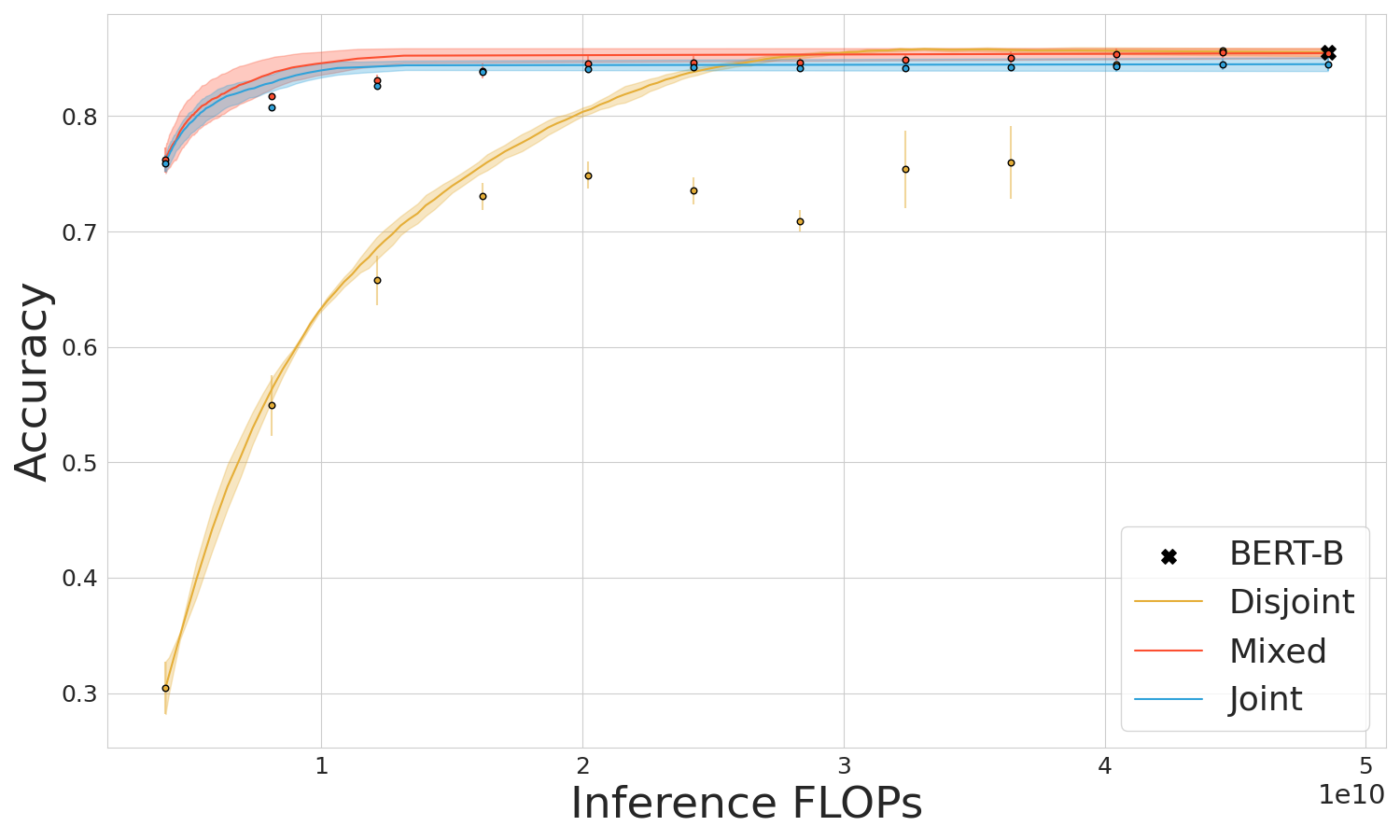}
        \caption{Performance–cost trade-off of the multi-exit BERT-B on the 20Newsgroups dataset}
    \end{minipage}
    \hfill
    \begin{minipage}{0.47\textwidth}
        \centering
        \includegraphics[width=\linewidth]{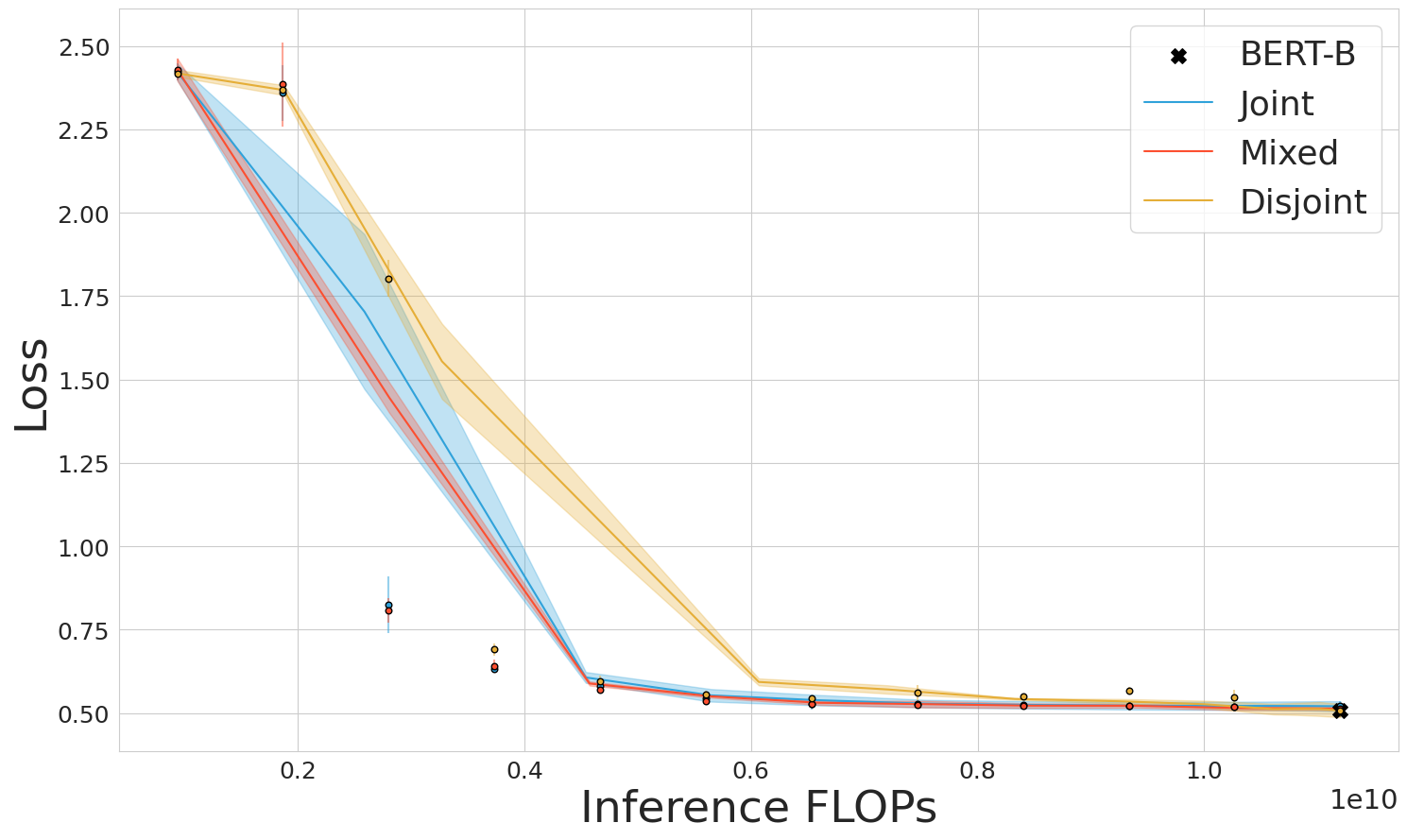}
        \caption{Performance–cost trade-off of the BERT-B architecture on the STS-B dataset}
    \end{minipage}

\end{figure}

\begin{figure}[h!]
    \centering

    % Row 1
    \begin{minipage}{0.48\textwidth}
        \centering
        \includegraphics[width=\linewidth]{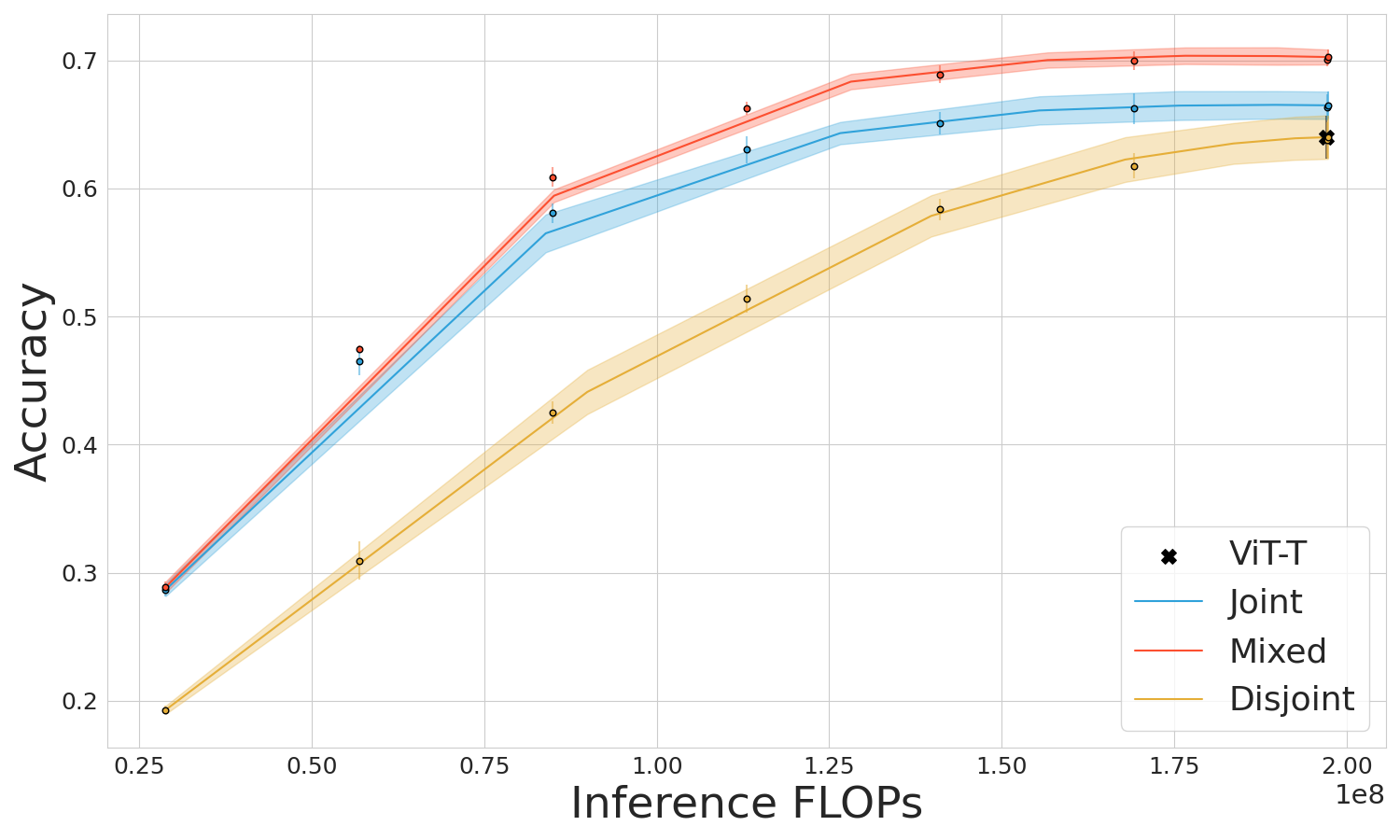}
        \caption{Performance–cost trade-off of the multi-exit PBEE with ViT-T as the backbone on the CIFAR-100 dataset}
    \end{minipage}
    \hfill
    \begin{minipage}{0.48\textwidth}
        \centering
        \includegraphics[width=\linewidth]{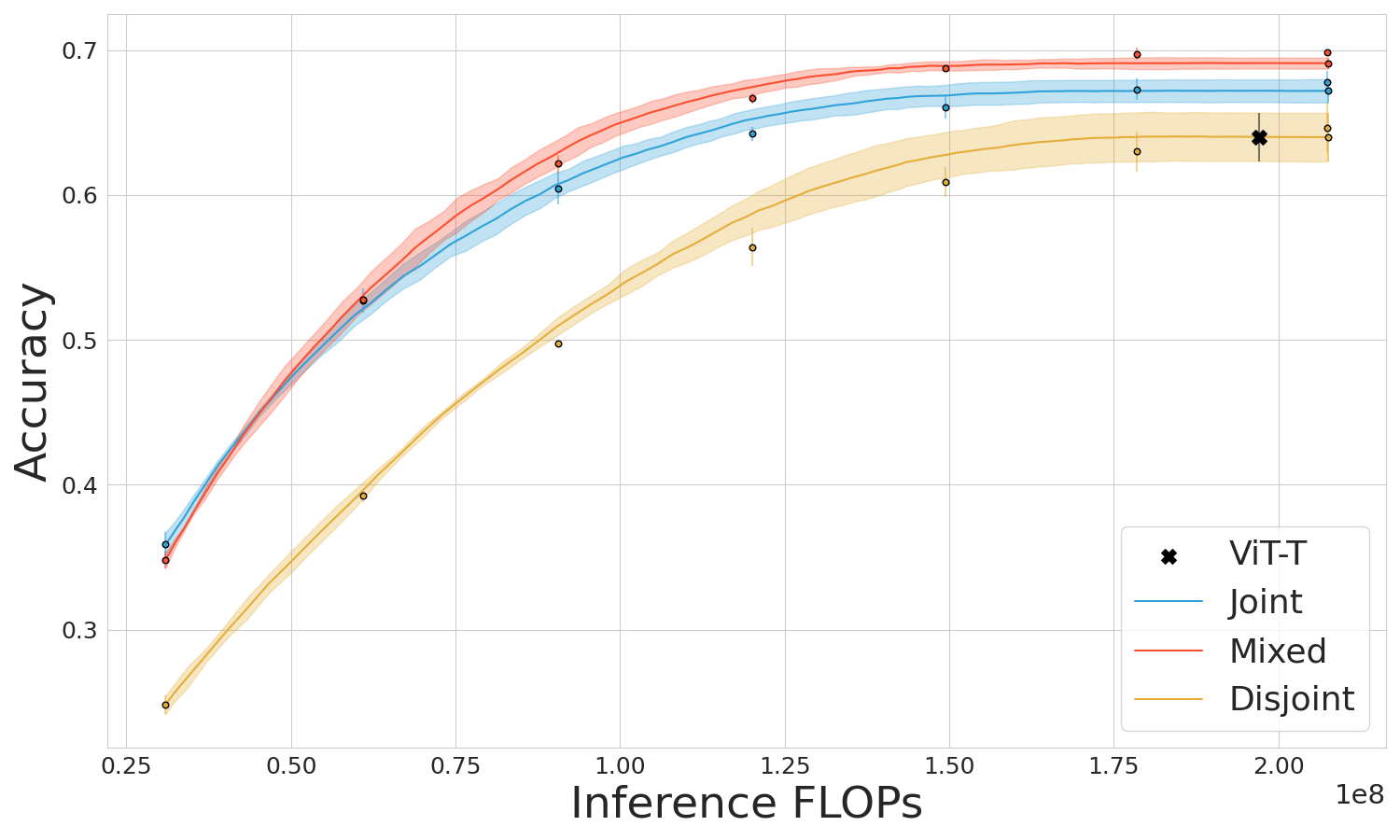}
        \caption{Performance–cost trade-off of the multi-exit GPF with ViT-T as the backbone on the CIFAR-100 dataset}
    \end{minipage}

    \vspace{1.2em}

    % Row 2
    \begin{minipage}{0.48\textwidth}
        \centering
        \includegraphics[width=\linewidth]{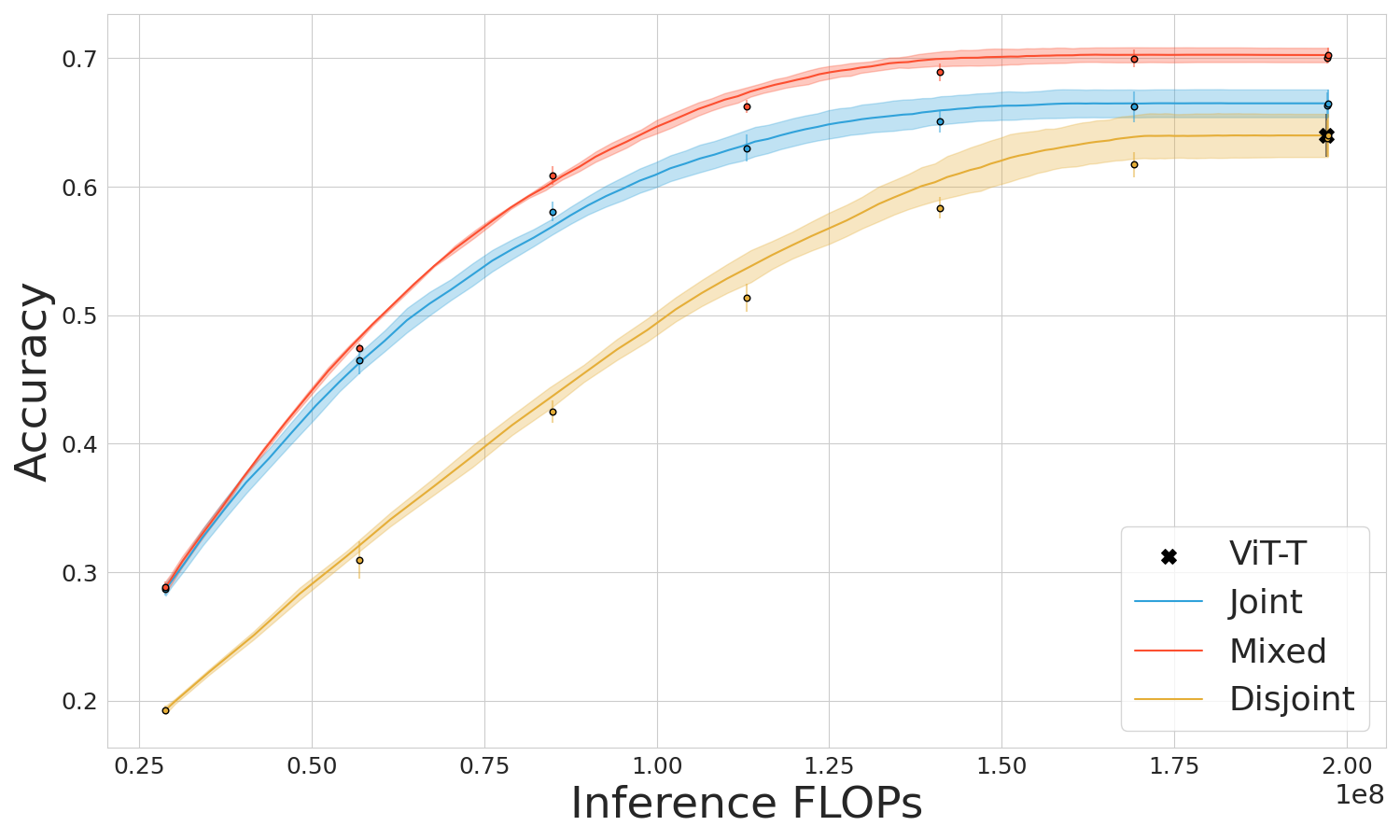}
        \caption{Performance–cost trade-off of the multi-exit ViT-T on the CIFAR-100 dataset, using normalized entropy as the exit criterion}
    \end{minipage}
    \hfill
    \begin{minipage}{0.48\textwidth}
        \centering
        \includegraphics[width=\linewidth]{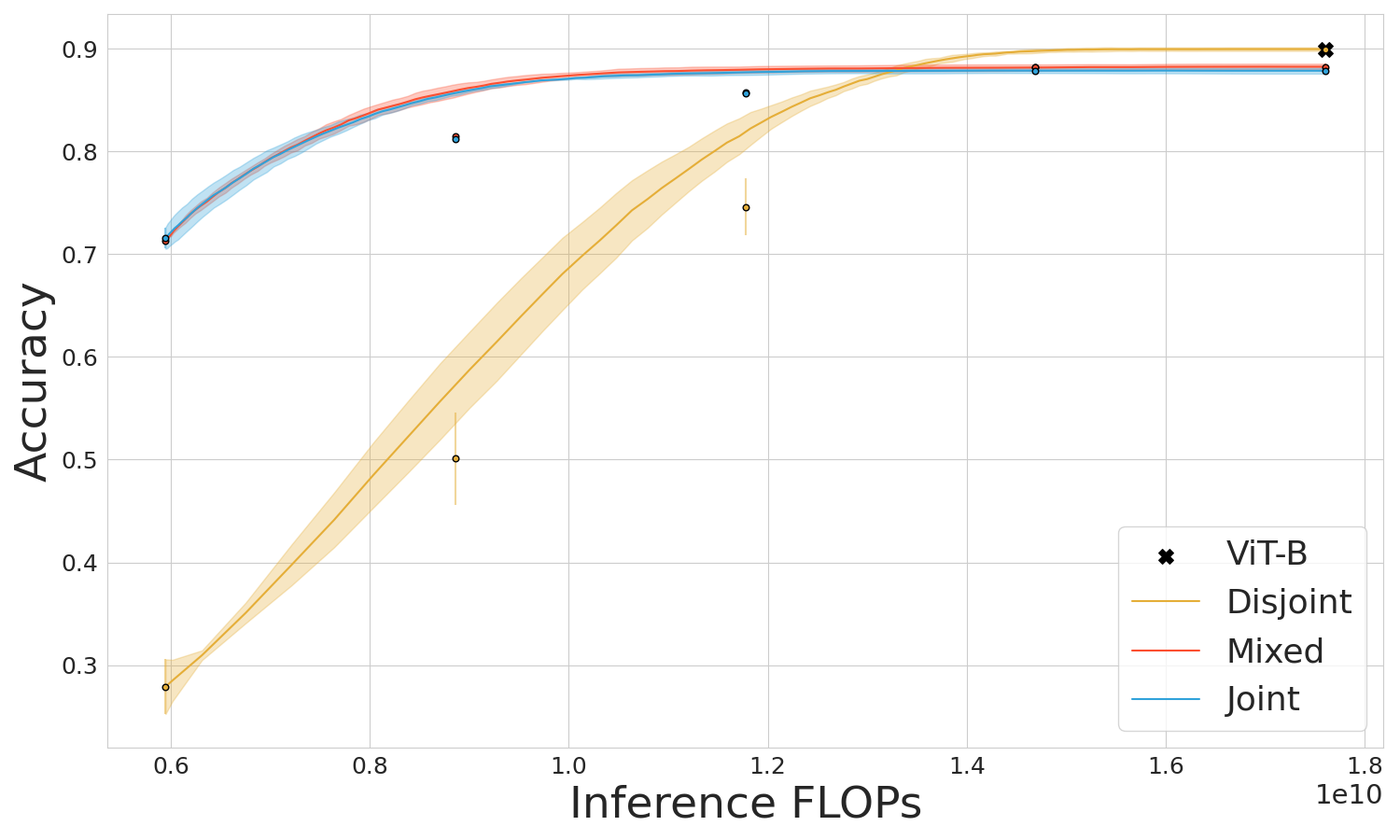}
        \caption{Performance–cost trade-off of the multi-exit ViT-B on the CIFAR-100 dataset, pre-trained on ImageNet-1k and fine-tuned on CIFAR-100}
    \end{minipage}

    \vspace{1.2em}

    % Row 3 – single centered image
    \hspace*{\fill}
    \begin{minipage}{0.48\textwidth}
        \centering
        \includegraphics[width=\linewidth]{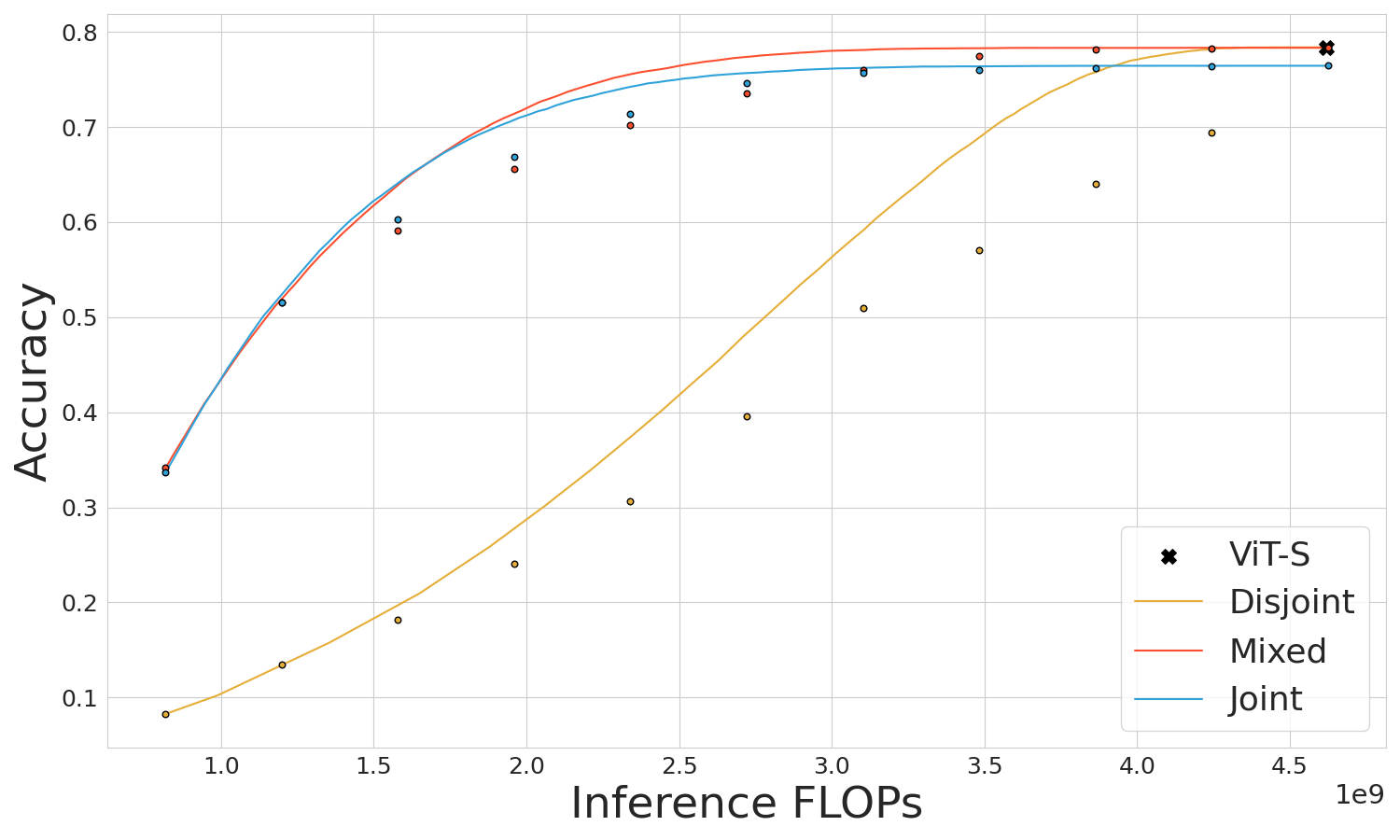}
        \caption{Performance–cost trade-off of the multi-exit ViT-S on the ImageNet-1k dataset}
    \end{minipage}
    \hspace*{\fill}

\end{figure}

\FloatBarrier
\section{Training Details}
\label{sec:training_details}
Here we provide more information training setup, in addition to what is described in~\Cref{par:exp_setup-train_setup}

\subsection{ResNet-34, CIFAR-100}
\paragraph{Model set-up.} 
Exit heads are placed at positions $\{2, 4, \dots, 14\}$.
\paragraph{Training set-up.} We train each model with batch size of 128. We use a learning rate of $5e{-4}$ and no weight decay. We set the early stopping patience to 50 epochs. CutMix and Mixup are used as augmentations.

\subsection{MSDNet, CIFAR-100}
\paragraph{Model set-up.}
We use CIFAR variant of MSDNet with 7 blocks. Exit heads are placed at positions $\{0, 1, \dots, 17\}$.
\paragraph{Training set-up} We train each model with batch size of 512. We use a learning rate  of $1e{-3}$ and no weight decay. We set the early stopping patience to 50 epochs. CutMix and Mixup are used as augmentations.

\subsection{ViT-T, CIFAR-100}
\paragraph{Model set-up.} ViT-T hyperparameters are as follows: a patch size of 4, an embedding size of 256, an MLP dimension of 256, 7 layers and 8 attention heads. Exit heads are placed at positions $\{0, 1, \dots, 6\}$. For GPF we set head embedding size to 512.
\paragraph{Training set-up} We train each model with batch size of 256. We use a learning rate  of $5e{-4}$ and no weight decay. We set the early stopping patience to 30 epochs. Following augmentations are used: random resizing, cropping, rotation, contrast adjustment, random erasing, CutMix and Mixup.

\subsection{ResNet-50, Tinyimagenet}
\paragraph{Model set-up.} Exit heads are placed at block positions $\{1..14\}$.
\paragraph{Training set-up.} We train each model with batch size of 256. We use a learning rate of $1e{-3}$ and no weight decay. We set the early stopping patience to 50 epochs. CutMix and Mixup are used as augmentations.

\subsection{ViT-T, ImageNet-1k}
\paragraph{Model set-up.}
ViT-T hyperparameters are as follows: a patch size of 16, an embedding size of 192, an MLP dimension of 768, 12 layers, and 3 attention heads. Exit heads are placed at positions $\{1, 2, \dots, 10\}$.
\paragraph{Training set-up.} We train each model using 4 A-100 GPUs with an effective batch size of 2048. We use a learning rate  of $5e{-4}$ and no weight decay. We set the early stopping patience to 25 epochs. Following augmentations are used: random resizing, cropping, rotation, contrast adjustment, random erasing, CutMix \cite{yun2019cutmix} and Mixup \cite{zhang2018mixup}.

\subsection{20 Newsgroups, Bert-B}
\paragraph{Model set-up}
We set maximum sequence length to 512. Exit heads are placed at positions $\{1, 3, \dots, 21\}$
\paragraph{Training set-up} We train each model with batch size of 32. We use a learning rate of $5e{-5}$ and a weight decay of $1e{-2}$. We set the early stopping patience to 3 epochs.

\subsection{STS-B, Bert-B}
\paragraph{Model set-up}
We set maximum sequence length to 128. Exit heads are placed at positions $\{1, 3, \dots, 21\}$
\paragraph{Training set-up} We train each model with batch size of 16. We use a learning rate of $1e{-5}$ and a weight decay of $1e{-4}$. We set the early stopping patience to 3 epochs.

\subsection{ViT-B, CIFAR-100}
\paragraph{Model set-up.} ViT-B hyperparameters are as follows: a patch size of 16, an embedding size of 768, an MLP dimension of 3072, 12 layers and 12 attention heads. Exit heads are placed at positions $\{3, 5, \dots, 9\}$.
\paragraph{Training set-up} We use pretrained weights from \verb|torchvision|~\citep{torchvision2016}, trained on ImageNet-1k. To fine-tune, we use a learning rate  of $2e{-5}$ and a weight decay of $3e{-2}$. We set the early stopping patience to 30 epochs. We use Mixup as an augmentation.

\end{document}